%% file: main.tex
\documentclass[11pt]{article}
\usepackage[preprint]{acl}

\usepackage{times}
\usepackage{latexsym}
\usepackage[T1]{fontenc}
\usepackage[utf8]{inputenc}
\usepackage{microtype}
\usepackage{inconsolata}

\usepackage{xurl}
\usepackage[htt]{hyphenat}
\usepackage{enumitem}

\usepackage{booktabs,tabularx,multirow,siunitx}
\usepackage{graphicx}
\usepackage{subcaption}
\usepackage{pgfplots}
\pgfplotsset{compat=1.18}

\usepackage{listings}
\usepackage[most]{tcolorbox}

\lstdefinelanguage{yaml}{
  keywords={true,false,null,y,n},
  keywordstyle=\color{blue}\bfseries,
  basicstyle=\ttfamily\footnotesize,
  sensitive=false,
  comment=[l]{\#},
  commentstyle=\color{gray},
  stringstyle=\color{teal},
  morestring=[b]',
  morestring=[b]"
}
\usepackage{pifont}
\newcommand{\cmark}{\ding{51}}
\newcommand{\xmark}{\ding{55}}

\usepackage{amsmath,amssymb}
\usepackage{algorithm}
\usepackage{algpseudocode}

\title{Executable Schema Contracts: From Automatic Ingestion to Multi-Source Retrieval}

\author{
Padmaja Jonnalagedda \quad Yuguang Yao \quad Xiang Gao \quad Hilaf Hasson\thanks{Work done while at Intuit AI Research. Now at Cohesity.} \quad Kamalika Das \\
Intuit AI Research \\
\texttt{saisri\_jonnalagedda@intuit.com}}

\begin{document}
\maketitle

\begin{abstract}
Real-world data spans tables, documents, and semi-structured files with implicit semantics. Querying this data requires integrating evidence across inconsistent schemas and formats, yet existing approaches either demand costly manual engineering or bypass structure entirely. We present a system that automatically discovers an executable schema from raw multi-source data and uses it as a shared contract for knowledge graph construction and query-time retrieval. A closed-world field catalog constrains LLM-based schema discovery to attested fields; deterministic structural analysis infers identity keys, foreign keys, and source hierarchy; and the resulting schema drives extraction, deduplication, and cross-source linking into a provenance-aware knowledge graph. At query time the schema -- optionally extended via a monotonic protocol -- conditions a multi-tool agent routing retrieval across structured lookup, graph traversal, and vector search, returning grounded answers with traceable citations. In controlled zero-shot comparisons using the same LLM, data, and evaluation harness, the system improves over retrieval-only and decomposition-based baselines across four QA benchmarks, with ablations showing that schema-conditioned routing, structural intelligence, and schema-guided construction each contribute to the gains.
\end{abstract}
\section{Introduction}
\label{sec:intro}
Practical knowledge rarely lives in a single database: it spans PDFs, tables, semi-structured JSONs, internal wikis, and ad-hoc spreadsheets with inconsistent identifiers and implicit semantics \cite{chen2020hybridqa}. Organizations increasingly rely on Question Answering (QA) systems to extract value from this mix (e.g., ``Which products launched after Company X's 2018 acquisitions, and what were their profits that year?'').
While Retrieval-Augmented Generation (RAG)~\cite{lewis2020rag} works for single-document lookup, it often fails on \emph{cross-source joins}, \emph{multi-hop retrieval}, and workflows requiring \emph{auditable provenance} \cite{barnett2024seven,liu2025hoprag,phanse2025msrs}. In practice, the hardest failures arise not from language modeling but from missing structure: systems cannot reliably align entities, traverse relationships, or justify answers across sources.

\textbf{The data-to-answer automation problem.}
We study self-configuring, self-extending QA over heterogeneous, multi-source data \emph{without manual schema design or ingestion}. This setting is non-stationary: sources change, identifiers drift, and new entity types emerge, yet QA must remain fast enough for interactive use, controllable, and auditable. Four bottlenecks recur:
\textbf{(1)} domain experts must predefine entity types, relationships, and extraction rules per dataset, repeating this as sources evolve;
\textbf{(2)} static schemas become outdated while fully open-ended approaches sacrifice correctness or latency;
\textbf{(3)} answers must be traceable to source records, requiring lineage throughout extraction, linking, and retrieval;
\textbf{(4)} multi-source questions require choosing retrieval strategies and bridging sources via joinable identifiers -- without explicit structure, routing is brittle.

\noindent Schema-light systems, which pair a large language model (LLM) with vector retrieval, reduce upfront effort but fail on cross-source joins and provenance-backed multi-hop QA. Manually engineered knowledge graphs (KGs) are precise but do not scale under schema changes~\cite{edge2024graphrag}. Moreover, extraction, linking, and retrieval are often optimized in isolation, leaving end-to-end correctness under schema evolution open.

\textbf{Our approach.}
We automatically discover a unified schema from heterogeneous sources and treat it as an \emph{executable contract} shared by offline ingestion and online QA -- so the system can decide where to look, how to connect evidence, and when not to answer, extending the schema at query time when needed.

\begin{figure*}[h!]
    \centering
    \includegraphics[width=\linewidth, trim={0.1cm 0.1cm 0.1cm 0.1cm},clip]{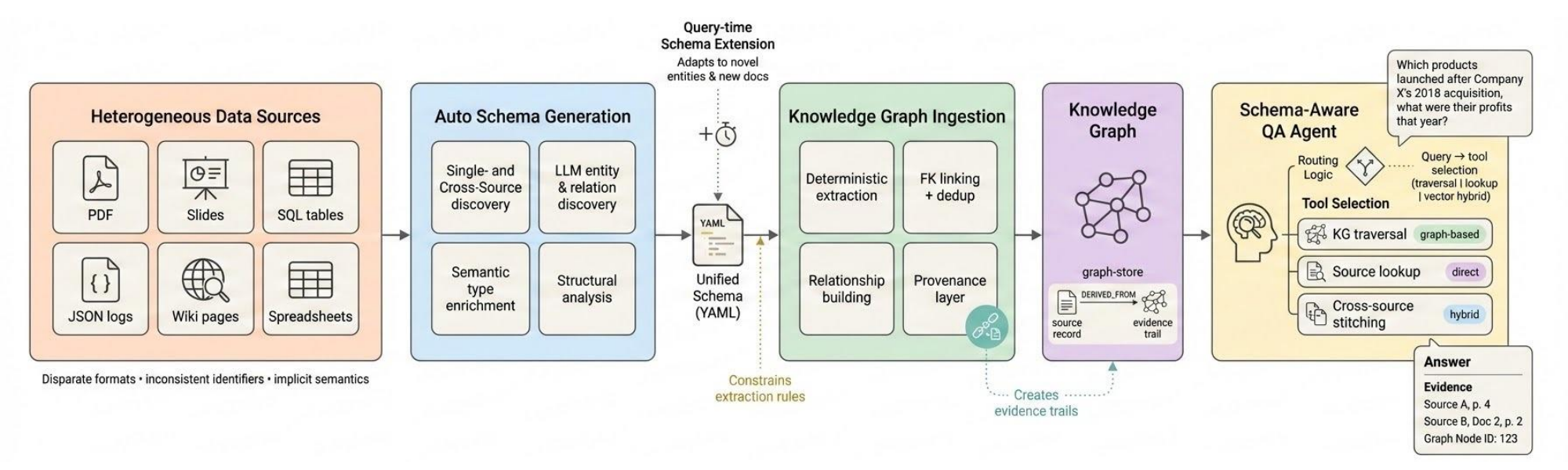}
    \vspace{-6mm}
    \caption{End-to-end pipeline.  Raw heterogeneous sources (tables, JSON, text) are profiled and fed to an automatic schema generator producing a closed-world field catalog and unified executable schema~$\Sigma$.  $\Sigma$ drives (i)~schema-guided KG ingestion -- extraction, deduplication, cross-source linking, provenance tracking -- into a typed Neo4j graph, and (ii)~a schema-guided QA agent routing queries across structured lookup, graph traversal, and vector retrieval, returning grounded answers with source citations.  Dashed paths indicate optional query-time extension.}
    \label{fig:overview}
\end{figure*}

\subsection{Contributions}
Our overarching contribution is an end-to-end formulation of heterogeneous QA around a shared \emph{executable schema contract}: one schema, induced from the data and restricted to observed fields, drives both KG construction and retrieval. We evaluate this formulation through three questions:

\begin{description}[leftmargin=0pt,itemsep=2pt]
    \item[\textbf{RQ1: Executable schema discovery.}]
    \emph{Can LLM-discovered schemas be made executable over raw heterogeneous sources?}
    A closed-world field catalog requires LLM-proposed attributes, extraction paths, and relationship link fields to reference attested field identifiers; invalid references are rejected or repaired before execution, turning schema discovery into a validated artifact that drives downstream extraction (Section~\ref{sec:schema-gen}).

    \item[\textbf{RQ2: Schema-guided construction.}]
    \emph{Does executing the induced schema during ingestion improve downstream QA?}
    The schema specifies extraction paths, identity keys, foreign-key links, and provenance hooks. Removing structural intelligence lowers exact match (EM) by 2.1--6.8 points and token-level F1 by up to 10.0 (metrics defined in Section~\ref{sec:metrics}, Section~\ref{sec:ablations}).

    \item[\textbf{RQ3: Schema-guided retrieval.}]
    \emph{Does exposing the induced schema to the QA agent improve retrieval over generic tool use?}
    Generic agents (ReAct, PlanAct) barely improve over RAG (+1.0--1.3 EM on BlendQA); the schema-guided agent improves by +10.2 EM. The full system achieves the best EM/F1 among controlled methods on all four benchmarks (Section~\ref{sec:main-results}).
\end{description}

\noindent We also study monotonic query-time schema extension as an optional mechanism for evolving schemas, and perform a cross-model evaluation on BlendQA with GPT-4.1, Claude Haiku~4.5, and Llama~3.3~70B, finding consistent gains across all three model families (Section~\ref{sec:model-agnostic-main}).

\section{Related Work}
\label{sec:related}
We study \emph{universal data-to-answer automation}: given heterogeneous sources, infer a schema, ingest data with lineage, and answer queries using the schema to decide \emph{where to look} and \emph{how to connect evidence}. This spans threads that typically address one pipeline layer or assume a fixed schema.

\noindent\textbf{(a) Schema discovery, KG construction, and entity resolution.}
Schema induction from text -- concept taxonomies \cite{wu2012probase}, lifelong extraction \cite{mitchell2018never} -- demonstrates the value of induced structure but remains text-centric. Recent work applies LLMs to entity matching and alignment across heterogeneous KGs \cite{wang2025entitymatching}. Our structural analysis connects to classic data profiling and schema matching; unlike these systems, we use the induced structure at run time to drive LLM-based ingestion and retrieval routing, targeting schema \emph{discovery} and cross-source \emph{linking} from raw data with link paths as routing primitives for multi-hop QA.

\noindent\textbf{(b) Natural language interfaces to structured and semi-structured data.}
Text-to-SQL \cite{yu2018spider,li2024dawn}, compositional table parsing \cite{pasupat2015compositional}, HybridQA \cite{chen2020hybridqa}, and the emerging Text-to-Cypher line \cite{ozsoy2025text2cypher,cazzaro2025spot} all assume a known, stable schema and focus on query generation given an existing substrate. Our focus is complementary: \emph{automating schema discovery, linking, and maintenance} as sources evolve.

\noindent\textbf{(c) Retrieval-augmented generation and graph-based retrieval.}
RAG \cite{lewis2020rag,gao2023rag_survey}, dense retrieval \cite{karpukhin2020dense}, hybrid retrieval \cite{chen2024bge}, structured KG retrieval \cite{sun2024thinkongraph}, and GraphRAG \cite{edge2024graphrag} all assume an existing index or engineered substrate. We instead \emph{discover retrieval structure from raw heterogeneous data} via schema induction, enabling downstream retrieval without repeated manual configuration.

\noindent\textbf{(d) Tool-use and agentic QA systems.}
Tool-using LLMs \cite{schick2023toolformer,qin2024toolllm} and prompting paradigms like ReAct \cite{yao2023react} and chain-of-thought \cite{wei2022chain} enable multi-step reasoning but leave tool and source selection to the model. We use the discovered schema as an \emph{agent contract} guiding routing among structured lookup, vector retrieval, and graph traversal.

\noindent\textbf{Closest comparisons.}
The three closest systems differ in what they assume as input.
\emph{AtomR}~\cite{xin2025atomr} decomposes questions into atomic operators over heterogeneous knowledge sources, but assumes those sources and their access interfaces already exist.
\emph{HippoRAG~2}~\cite{gutierrez2025rag} and \emph{GraphRAG}~\cite{edge2024graphrag} build graphs automatically, but from text passages and without a typed schema: HippoRAG~2 builds an open triple graph for memory-style retrieval, and GraphRAG builds community summaries over an LLM-extracted entity graph.
None of them induces a typed schema with identity and foreign keys across sources, or extends that schema at query time---which our system automates. Appendix~\ref{app:closest} compares the four systems side by side.

\begin{figure*}[t]
    \centering
    \includegraphics[width=0.93\linewidth, trim={0.1cm 0.2cm 0.1cm 0},clip]{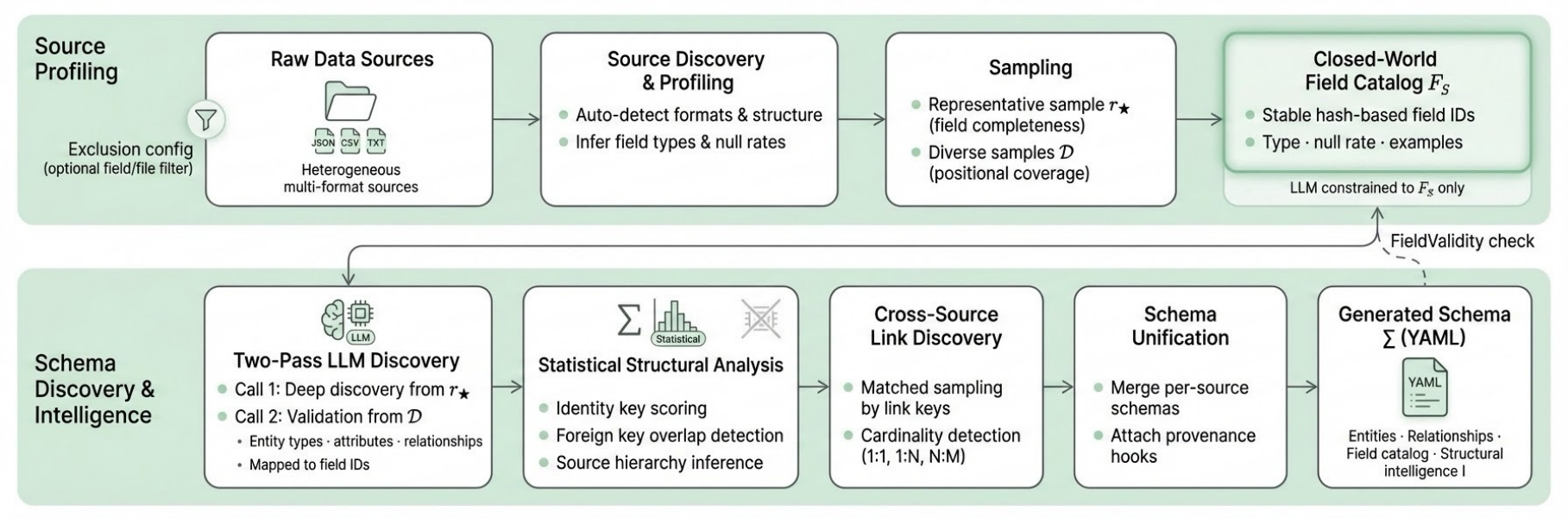}
    \vspace{-2mm}
    \caption{Automatic schema generation: closed-world field catalog $\rightarrow$ two-pass semantic discovery $\rightarrow$ structural/cross-source intelligence.}
    \label{fig:schemagen}
\end{figure*}

\section{Technical Approach}
\label{sec:technical}

Rather than treating profiling, extraction, KG construction, and retrieval as separate stages, we organize them around a single induced schema $\Sigma$ that acts as an \emph{executable contract} (Figure~\ref{fig:overview}): it determines what the LLM can reference and extract (closed-world catalog), how entities are linked (structural intelligence), and where the agent retrieves at query time (schema-conditioned routing). We define the contract (Section~\ref{sec:problem}), then describe how it is induced (Section~\ref{sec:schema-gen}), executed to build the KG (Section~\ref{sec:ingest-main}), extended at query time (Section~\ref{sec:schema-ext}), and used to route queries (Section~\ref{sec:agent-main}).

\subsection{Schema Contract Definition}
\label{sec:problem}

Let $\mathcal{S} = \{S_1, \dots, S_m\}$ be heterogeneous data sources (tables, JSON files, text corpora) with unknown and potentially overlapping schemas. Given a natural-language query $Q$, the goal is to return a grounded answer $\hat{y}$ with optional provenance traces to originating records. We decompose this into three sub-problems:

\begin{enumerate}
    \item \textbf{Schema Induction.} Discover a unified schema $\Sigma = (\mathcal{F}, \mathcal{T}, \mathcal{R}, \mathcal{I})$ where $\mathcal{F}$ is a closed-world field catalog (the set of fields observed in $\mathcal{S}$), $\mathcal{T}$ a set of entity types with attribute mappings $\mathcal{A}: t \mapsto 2^{\mathcal{F}}$ for each $t \in \mathcal{T}$, $\mathcal{R}$ relationship types, and $\mathcal{I}$ structural intelligence (identity keys, foreign keys (FKs), source hierarchy, extraction paths).
    \item \textbf{Schema-Guided Construction.} Using $\Sigma$ to specify extraction and linking, build a provenance-aware knowledge graph $G = (V, E, \pi)$ where $V$ contains typed entities, $E$ typed relationships, and $\pi: V \cup E \rightarrow 2^{\mathcal{S}}$ maps each element to its source records.
    \item \textbf{Schema-Guided QA.} Given $Q$, optionally extend $\Sigma$ to a schema $\Sigma' \supseteq \Sigma$ (Section~\ref{sec:schema-ext}; $\Sigma'=\Sigma$ when no extension is needed), then select and compose retrieval tools $\{$\textsc{SchemaLookup}, \textsc{GraphTraverse}, \textsc{VectorRAG}$\}$ using $\Sigma'$ to produce $\hat{y}$ with citations $\pi(\hat{y}) \subseteq \mathcal{S}$.
\end{enumerate}
\noindent $\Sigma$ is stored as a versioned YAML file and $G$ as a Neo4j property graph, with text chunked and indexed for dense retrieval (Appendix~\ref{app:schema:notation}).

\subsection{Inducing the Contract from Raw Sources}
\label{sec:schema-gen}

Given heterogeneous sources $\mathcal{S}$, schema induction must discover entity types, attribute mappings, and cross-source relationships without manual annotation.
Two challenges motivate our design: (1)~LLM-based discovery hallucinates fields absent from the data, producing non-executable schemas; (2)~structural patterns -- foreign keys, cardinality, identity keys -- are unreliable when inferred semantically.
We address both via \emph{closed-world grounding}, restricting all LLM outputs to attested data fields, and \emph{statistical-before-semantic} ordering, delegating structural inference to deterministic methods and reserving LLM calls for semantic discovery only. This pipeline is shown in Figure~\ref{fig:schemagen}. Pseudocode and details are provided in Appendix~\ref{app:schema}.

\noindent \textbf{(i) Closed-world constrained discovery.}
We profile each source $S \in \mathcal{S}$ and construct a \emph{field catalog} $\mathcal{F}_S$ listing every observable field with a stable hash-based ID, data type, null rate, and example values; the global catalog is $\mathcal{F} = \bigcup_{S} \mathcal{F}_S$. The LLM may reference only identifiers in $\mathcal{F}_S$; every output is validated via:
\begin{equation}
\label{eq:field-validity-main}
\mathit{FieldValidity} = {|\{\text{ref.\ IDs}\} \cap \mathcal{F}_S|}\;/\;{|\{\text{ref.\ IDs}\}|}
\end{equation}
Schemas with $\mathit{FieldValidity} < 1.0$ are flagged for repair. This prevents hallucinated \emph{field references}, though not incorrect entity types or spurious relationships among valid fields.

Semantic discovery uses two LLM calls per source (cross-source link detection may add more; Appendix~\ref{app:schema:crosslink}), balancing quality and cost.
\emph{Call~1 (deep discovery)} analyzes a representative sample $r^\star$, selected for field completeness, together with $\mathcal{F}_S$ to extract entity types, attributes mapped to field IDs, and relationships.
\emph{Call~2 (validation)} reviews diverse samples $\mathcal{D}$ drawn from different positions to surface entities or relationships absent from $r^\star$.
The outputs are merged and a deterministic validator checks that every field reference exists in $\mathcal{F}_S$ and every relationship connects defined entity types (Appendix~\ref{app:schema:llm:merge}). The second pass recovers entity types that one pass misses, especially rare types in non-representative regions.

\noindent \textbf{(ii) Statistical structural intelligence.}
Uniqueness, cardinality, and value overlap are exact statistics that data profiling computes directly~\cite{abedjan2015profiling}, whereas LLMs remain unreliable even at simple structural and counting tasks such as table size or occurrence counts~\cite{sui2024table,xu2025genius,chang2025count,fu2024count}. We therefore compute them deterministically, without the LLM; removing it lowers EM by 2.1--6.8 points (Section~\ref{sec:ablations}).
\begin{itemize}[leftmargin=*,nosep]
  \item \textit{Identity keys.} A field qualifies if its uniqueness ratio is ${\geq}0.95$, or ${\geq}0.80$ with an ID-like name (Eq.~\ref{eq:id-gate}, Appendix~\ref{app:schema:struct}). They serve as deterministic dedup keys in ingestion, resolving duplicates via $O(n)$ hashing when keys exist.
  \item \textit{Foreign keys (FKs).} For each source pair, we compute field-level value overlap (Eq.~\ref{eq:fk-overlap}) and emit column pairs with non-zero overlap as \emph{candidate} join keys with confidence $\min\bigl(0.95,\; 0.5 + 0.3\,\mathit{overlap} + 0.15\,\mathbf{1}_{\text{name}}\bigr)$, where $\mathbf{1}_{\text{name}}$ indicates identical column names. Low-overlap candidates thus receive lower confidence. Cardinality (1:1, 1:N, N:M) is inferred from records per shared key value.
  \item \textit{Source hierarchy.} In a 1:N FK relationship, the ``1'' side (fewer records per key) is the parent.
\end{itemize}
\noindent These patterns are encoded in $\mathcal{I}$ and used downstream for deduplication, cross-source linking, and extraction-path selection.

\noindent \textbf{(iii) Cross-source link discovery and unification.}
Naive parallel indexing -- comparing the $i$-th record from source $A$ with the $i$-th from $B$ -- fails for 1:N and N:M relationships. We instead apply \emph{matched sampling}: for each shared value of a candidate link key (a field with high value overlap), all matching records from both sources are analyzed as a group. This handles arbitrary cardinalities. Per-source schemas are unified into a global $\Sigma$ with provenance hooks tracking each entity's origin.

\noindent An optional exclusion manifest prevents ground-truth leakage in evaluation and redacts sensitive fields in production (Appendix~\ref{app:schema}).

\subsection{Executing the Contract: KG Construction}
\label{sec:ingest-main}

Given $\Sigma$, we build a typed, provenance-aware Neo4j KG. The key property is that $\Sigma$ alone determines \emph{what} is extracted, \emph{how} duplicates are resolved, and \emph{where} cross-source links are created -- changing the schema changes the KG without code changes (Figure~\ref{fig:kgingest}, Appendix~\ref{app:ingest}).

\noindent \textbf{Schema-driven extraction.}
$\Sigma$'s attribute mappings $\mathcal{A}(t)$ select the extraction strategy per source: deterministic extractors handle schema-mapped fields; LLM extraction fires only when $\Sigma$ marks a source as hybrid. Entities are validated against $\Sigma$'s type constraints.

\noindent \textbf{Schema-driven deduplication.}
$\Sigma$'s identity keys ($\mathcal{I}$) set the dedup strategy: types with keys use exact key grouping ($O(n)$); others fall back to embedding clustering (cosine similarity ${\geq}0.85$; thresholds and rationale in Appendix~\ref{app:ingest:thresholds}).

\noindent \textbf{Schema-driven linking.}
Cross-source edges are created where $\Sigma$ specifies foreign keys: hash indexes over FK columns materialize edges for matching values, enabling cross-source multi-hop queries. Within-source relationships use schema link fields and record co-occurrence.

\noindent \textbf{Provenance.}
Each raw record becomes a \texttt{SourceRecord} node with \texttt{DERIVED\_FROM} edges, so agent citations are traceable (Section~\ref{sec:agent-main}).

\begin{figure*}[t]
    \centering
    \includegraphics[width=0.93\linewidth]{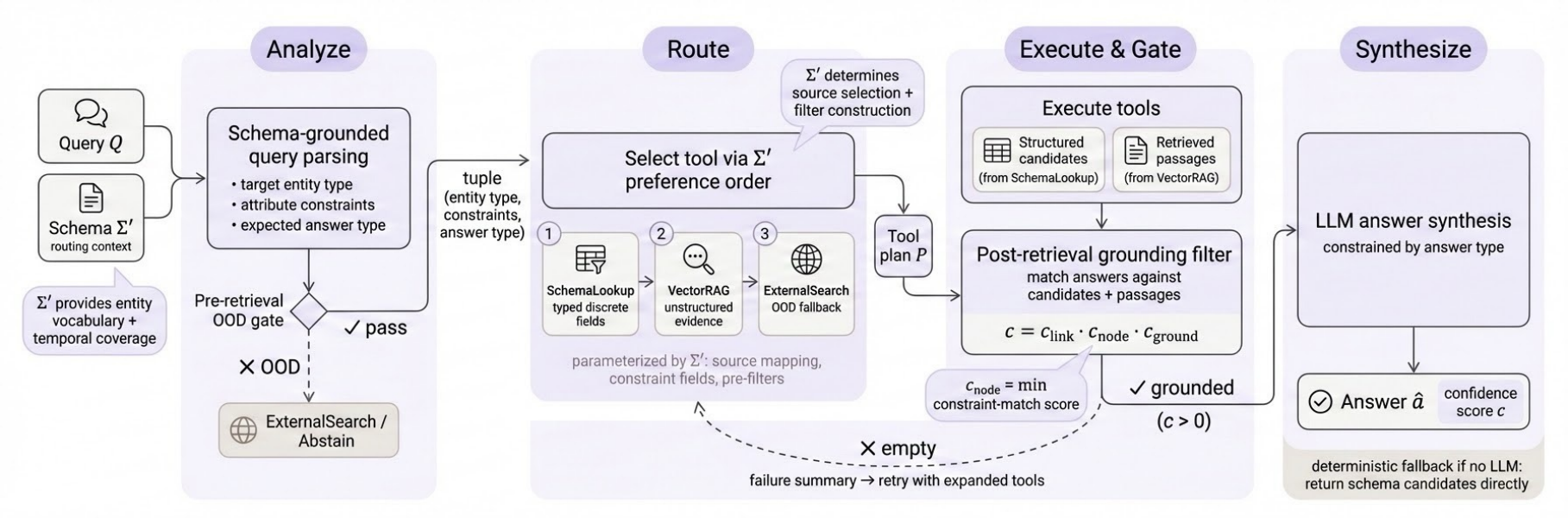}
    \vspace{-2mm}
    \caption{Schema-guided QA agent. The agent parses entity types and constraints, routes via $\Sigma'$'s preference order, applies pre-/post-retrieval reliability gating ($c = c_{\text{link}} \cdot c_{\text{node}} \cdot c_{\text{ground}}$), and synthesizes the grounded answer.}
    \label{fig:fig4_agent}
\end{figure*}

\subsection{Extending the Contract at Query Time}
\label{sec:schema-ext}

A static $\Sigma$ cannot anticipate every query. We formalize \emph{conditional schema augmentation}: given a trigger $T$ (query or document), determine whether $\Sigma$ suffices; if not, produce $\Sigma' \supseteq \Sigma$ under a \emph{monotonicity invariant} ($\mathcal{T}' \supseteq \mathcal{T}$, $\mathcal{R}' \supseteq \mathcal{R}$, $\mathcal{A}'(t) \supseteq \mathcal{A}(t)$), so prior schema elements remain valid.

\noindent \textbf{Sufficiency gating.}
A lightweight check extracts entity types from $T$ and verifies them against $\Sigma$'s vocabulary; extension fires only on unresolved types (Algorithm~\ref{alg:query-sufficiency}, Appendix~\ref{app:extend:sufficiency}). Extension is thus needed only for types absent from $\Sigma$; otherwise the static schema plus fallback retrieval suffices. On TAT-QA the gate triggers on 11.1\% of queries; on HybridQA, where each query has its own knowledge base (KB), it triggers on 85.3\%, and in 98.4\% of triggered cases at least one added type is used in retrieval (Section~\ref{sec:online-perf}).

\noindent \textbf{Diff-based merge.}
When triggered, the system profiles the new context, extracts candidate schema elements, and computes a schema diff. Query-scoped additions need confidence ${\geq}0.7$ to be merged; existing definitions are never altered.

\noindent \textbf{Two modes.}
\emph{Persistent extension} handles corpus growth via the full two-pass pipeline (Section~\ref{sec:schema-gen}), saving $\Sigma'$.
\emph{Query-scoped extension} handles concept gaps via one LLM call yielding an in-memory $\Sigma'$ discarded after the query (Appendix~\ref{app:extend}).

\subsection{Executing the Contract: Query Routing}
\label{sec:agent-main}

Given query $Q$, schema $\Sigma'$ (possibly extended), and heterogeneous retrieval backends -- structured stores and a dense text index -- the agent must select and compose retrieval tools to produce a grounded answer~$\hat{y}$.
Standard tool-use agents (e.g., ReAct~\cite{yao2023react}) rely on LLM reasoning to select tools at each step; this is expensive and blind to corpus structure.
We contribute \textbf{schema-conditioned routing} -- tool selection and query planning derived from the schema -- together with \textbf{two-stage reliability gating} for abstention (Figure~\ref{fig:fig4_agent}; Appendix~\ref{app:agent}).

\noindent \textbf{(i) Schema-conditioned routing.}
At initialization the agent introspects $\Sigma'$ to build a runtime \emph{routing context}: (a)~a field-type partition (typed columns with discrete values vs.\ free text), (b)~entity and relationship vocabulary, and (c)~known-value dictionaries from field-level statistics in $\Sigma'$.
At query time it parses $Q$ against this context to identify target entity types, attribute constraints, and expected answer type -- all anchored in $\Sigma'$'s vocabulary (Appendix~\ref{app:agent:analysis}).
Routing follows a preference order: \textsc{SchemaLookup} for typed discrete-valued constraints, \textsc{VectorRAG} for unstructured evidence, \textsc{GraphTraverse} for multi-hop and relationship queries over the KG, and \textsc{ExternalSearch} (disabled in our experiments) for out-of-distribution (OOD) queries that fall outside $\Sigma'$. Each tool's behavior is parameterized by $\Sigma'$ -- source selection, constraint fields, and filter construction are derived from the schema, not hard-coded (Appendix~\ref{app:agent:tools}).

\noindent \textbf{(ii) Two-stage reliability gating.}
Rather than answering every query, the agent performs two reliability checks.
\emph{Pre-retrieval} (Stage~1): a schema-coverage gate verifies that $Q$'s entity mentions and temporal references fall within $\Sigma'$'s known value sets and temporal range; queries that fail are flagged OOD and routed to abstention (Appendix~\ref{app:agent:ood}).
\emph{Post-retrieval} (Stage~2): a grounding filter checks whether candidate answers are supported by retrieved evidence.
Composite confidence $c = c_{\text{link}}\cdot c_{\text{node}}\cdot c_{\text{ground}}$ aggregates path confidence, constraint-match quality, and grounding strength.
If no answer is grounded, a structured \emph{failure summary} triggers retry (Appendix~\ref{app:agent:loop}).

\noindent \textbf{(iii) Synthesis.}
Evidence from all tools is merged into one context for LLM answer synthesis (fallback behavior and traces in Appendix~\ref{app:agent:traces}).

\section{Experiments and Results}
\label{sec:experiments-results}

We evaluate end-to-end QA accuracy across four benchmarks (Table~\ref{tab:datasets}), intrinsic schema/KG quality, and online serving efficiency.
All methods operate over identical dataset-provided sources.

\subsection{Experimental Setup}
\label{sec:setup}

\subsubsection{Benchmarks.}
Existing QA benchmarks~\cite{rajpurkar2016squad,yang2018hotpotqa,ho2020constructing,wu2024stark,yang2024crag} assume pre-curated corpora, an existing KB, or focus on online retrieval.
We select four benchmarks (Table~\ref{tab:datasets}) covering complementary dimensions:
\textbf{\textsc{BlendQA}}~\cite{xin2025atomr} stresses cross-source discovery and routing under heterogeneity (text + structured facts, multi-hop comparisons).
\textbf{\textsc{HybridQA}}~\cite{chen2020hybridqa} requires bridging table rows and linked passages; each query has its own KB, requiring per-query extension.
\textbf{\textsc{TAT-QA}}~\cite{zhu2021tatqa} covers financial QA over tables and passages with numeric reasoning; it occasionally requires extension.
\textbf{\textsc{ComplexTR}}~\cite{tan2023complextr} tests temporal multi-hop reasoning over time-scoped fact sentences.
All datasets are publicly available for research use. 

\begin{table}[t]
\small
\centering
\setlength{\tabcolsep}{3pt}
\scalebox{0.78}{%
\begin{tabular}{lrllccc}
\toprule
\textbf{Dataset} & \textbf{\#Q} & \textbf{Source types} & \textbf{MS} & \textbf{MH} & \textbf{Temp.} & \textbf{Ext.} \\
\midrule
\textsc{BlendQA}   & 445  & Wikidata, passages, facts & $\checkmark$ & $\checkmark$ & $\times$      & $\times$ \\
\textsc{HybridQA}  & 3{,}466 & Tables + Wikipedia passages & $\checkmark$ & $\checkmark$ & $\times$      & $\checkmark$ \\
\textsc{TAT-QA}    & 1{,}668 & Financial tables + text & $\checkmark$ & $\times$     & $\times$  & $\sim$ \\
\textsc{ComplexTR} & 1{,}000 & Temporal fact sentences & $\times$ & $\checkmark$ & $\checkmark$  & $\times$ \\
\bottomrule
\end{tabular}%
}
\caption{Dataset characteristics. $\checkmark$/$\times$/$\sim$ = yes/no/sometimes. MS = multi-source, MH = multi-hop, Temp.\ = temporal, Ext.\ = needs query-time extension.}
\label{tab:datasets}
\vspace{-3mm}
\end{table}

\begin{table*}[t]
\centering
\footnotesize
\setlength{\tabcolsep}{2.8pt}
\begin{tabular}{l|cccc|cccc|cccc|cccc}
\toprule
\multirow{2}{*}{\textbf{Method}} &
\multicolumn{4}{c|}{\textbf{\textsc{BlendQA}}} &
\multicolumn{4}{c|}{\textbf{\textsc{HybridQA}}} &
\multicolumn{4}{c|}{\textbf{\textsc{TAT-QA}}} &
\multicolumn{4}{c}{\textbf{\textsc{ComplexTR}}} \\
& \textbf{EM} & \textbf{F1} & \textbf{BLEU} & \textbf{LJ}
& \textbf{EM} & \textbf{F1} & \textbf{BLEU} & \textbf{LJ}
& \textbf{EM} & \textbf{F1} & \textbf{BLEU} & \textbf{LJ}
& \textbf{EM} & \textbf{F1} & \textbf{BLEU} & \textbf{LJ} \\
\midrule
\multicolumn{17}{l}{\textit{Controlled re-runs (same GPT-4.1 backbone, sources, exclusions, and harness)}} \\
RAG & 19.1 & 33.6 & 26.1 & 51.0 & 7.0 & 13.8 & 14.2 & 17.5 & 18.5 & 33.9 & 24.9 & 41.3 & 25.5 & 53.3 & 44.6 & 66.5 \\
ProbTree & 20.1 & 33.9 & 36.6 & 52.4 & 55.5 & 65.9 & 66.4 & 74.2 & 33.2 & 48.0 & 48.4 & 64.7 & 70.0 & 80.5 & 79.7 & 73.6 \\
CoK & 20.4 & 38.6 & 40.5 & 59.2 & 37.5 & 45.9 & 45.8 & 49.8 & 26.6 & 33.0 & 33.9 & 44.3 & 28.0 & 40.0 & 38.6 & 44.2 \\
AtomR & 26.7 & 43.3 & 49.2 & 63.4 & 38.5 & 50.4 & 49.1 & 60.3 & 68.6 & 74.9 & 62.1 & 85.6 & 64.0 & 75.7 & 74.7 & 74.1 \\
\midrule
\multicolumn{17}{l}{\textit{Best published result (not controlled; for context only)$^\ddagger$}} \\
Best published &  -- & 43.3  & -- & -- & 51.5 & 66.0  & -- & -- & 71.9  & 79.7 & -- & -- & -- & 78.2 & 72.7  & 81.5 \\
\midrule
\multicolumn{17}{l}{\textit{Ours}} \\
\textbf{Schema-Guided} & \textbf{29.3} & \textbf{48.9} & \textbf{54.1} & \textbf{70.0} & \textbf{58.3} & \textbf{68.9} & \textbf{69.0} & \textbf{78.5} & \textbf{74.3} & \textbf{82.6} & \textbf{68.6} & \textbf{91.8} & \textbf{74.5} & \textbf{82.2} & \textbf{82.0} & \textbf{76.3} \\
\midrule
\multicolumn{17}{l}{\textit{Ablations (each removes one component, all else fixed)}} \\
w/o schema guidance & 20.0 & 32.3 & 39.3 & 52.1 & 55.6 & 68.0 & 66.9 & 78.0 & 72.2 & 77.9 & 66.2 & 89.1 & 67.2 & 77.0 & 76.8 & 72.5 \\
w/o KG ingestion &  21.3 & 36.5 & 43.1 & 57.4 & 57.8 & 68.4 & 68.4 & 78.0 & 71.7 & 77.3 & 66.8 & 89.6 & 68.8 & 77.1 & 76.8 & 73.8 \\
single-pass & 25.4 & 41.8 & 47.9 & 63.3 & 57.8 & 67.7 & 67.9 & 77.2 & 71.1 & 76.2 & 65.3 & 86.2 & 65.0 & 76.3 & 75.7 & 71.6 \\
w/o schema exten. & N/A & N/A & N/A & N/A & 55.5 & 66.0 & 66.2 & 75.1 & 72.8 & 78.6 & 67.0 & 90.4 & N/A & N/A & N/A & N/A \\
w/o struc. intel. & 26.3 & 38.9 & 44.8 & 54.5 & 55.0 & 68.1 & 67.2 & 76.0 & 72.2 & 76.9 & 66.7 & 89.9 & 67.7 & 78.6 & 77.9 & 74.9 \\
\bottomrule
\multicolumn{17}{l}{\footnotesize $^\ddagger$Best published zero-shot / no-finetuning references: BlendQA~\cite{xin2025atomr}, HybridQA~\cite{agarwal2025hybrid},} \\
\multicolumn{17}{l}{\footnotesize \ \ 
 TAT-QA GPT-4 zero-shot from \citet{zhu2024tat}, ComplexTR HippoRAG~2 as reported in \citet{shu2026remem}.}
\end{tabular}
\caption{End-to-end QA performance (EM, F1, BLEU, and LLM-judge accuracy LJ; Section~\ref{sec:metrics}). \emph{Controlled re-runs} are our implementations with the same GPT-4.1 backbone, dataset-provided sources, ground-truth exclusion manifest, evaluation harness, and zero-shot setting as our method. \emph{Best published result} is the strongest previously reported no-finetuning result per benchmark that we could identify; it is not controlled for model, prompting, retrieval substrate, or implementation and is not the basis of our claims. \emph{Ablations} remove the specified component while keeping the remaining infrastructure fixed.}
\label{tab:qa_all_metrics}
\end{table*}

\subsubsection{Baselines.}
We compare against: (i)~\textbf{RAG}~\cite{lewis2020rag}, a retrieve-then-generate baseline using the top-$k$ passages; (ii)~\textbf{ProbTree}~\cite{cao2023probabilistic}, organizing multi-hop retrieval as probabilistic tree search over evidence chains; (iii)~\textbf{CoK}~\cite{li2023chain}, iteratively gathering and verifying facts via chain-of-knowledge decomposition; and (iv)~\textbf{AtomR}~\cite{xin2025atomr}, decomposing reasoning into atomic operator calls over structured and textual evidence.
For context, we also report the \textbf{best published result} per benchmark, i.e., the strongest zero-shot / no-finetuning number we could find in the literature: AtomR~\cite{xin2025atomr} (BlendQA), Odyssey~\cite{agarwal2025hybrid} (HybridQA), GPT-4 zero-shot from \citet{zhu2024tat} (TAT-QA), and HippoRAG~2~\cite{gutierrez2025rag} as reported in \citet{shu2026remem} (ComplexTR), copied from the cited papers for context only.

\paragraph{Baseline fidelity.}
Two anchors indicate our controlled re-runs are not weakened versions of the original methods. On BlendQA, where AtomR was introduced, our AtomR re-run obtains 43.3 F1, matching the published result. On HybridQA, our ProbTree re-run (55.5 EM) exceeds the best published result (51.5 EM). AtomR's larger gap on HybridQA (38.5 EM) reflects a setting mismatch, as AtomR was not designed for per-query KBs. 

\paragraph{Models and configuration.}
All experiments are \textbf{zero-shot} with GPT-4.1 (temperature~$=0$); all controlled baselines use the same backbone. The QA agent uses E5 bi-encoder embeddings with FAISS and optional cross-encoder reranking. \textsc{ExternalSearch} is disabled in benchmark runs, so OOD queries are routed to abstention and all results use only dataset-provided sources. Details, pseudocode, and prompts are in the Appendix; code will be released upon acceptance. We also report results with other LLM backbones (Section~\ref{sec:model-agnostic-main}, Appendix~\ref{app:crossmodel-tatqa}) and over three runs on BlendQA (Appendix~\ref{app:variance}).

\subsubsection{Evaluation metrics.}
\label{sec:metrics}
We report four metrics, all as percentages. \textbf{Exact match (EM)}~\cite{rajpurkar2016squad} is the fraction of predictions identical to the gold answer after normalization (lower-casing, punctuation/article removal). \textbf{F1}~\cite{rajpurkar2016squad} is the harmonic mean of token-level precision and recall between prediction and gold answer. \textbf{BLEU}~\cite{papineni2002bleu} measures n-gram precision of the prediction against the gold answer. \textbf{LLM-J} (LJ in tables) is LLM-as-judge accuracy~\cite{zheng2023judging}: Claude Haiku~4.5 at temperature~$=0$ receives the predicted and gold answers and returns 1 if they are semantically equivalent and 0 otherwise.
EM and F1 are the primary metrics and do not depend on a judge; LLM-J captures semantic equivalence beyond string overlap. A BlendQA rescore with three judges from different model families gives similar accuracy (Appendix~\ref{app:multijudge}).
Official evaluation protocols are used if available (e.g., TAT-QA).
\paragraph{Evaluation hygiene.}
For all datasets, annotation fields and files are removed before schema induction, KG construction, vector indexing, and query-time extension (Table~\ref{tab:leakage-control}). In TAT-QA, this removes the \texttt{questions} subtree; we index only the table and paragraph evidence available to all methods. Appendix~\ref{app:schema:audit} gives a stage-by-stage audit.

\subsection{Results}
\label{sec:main-results}

Table~\ref{tab:qa_all_metrics} summarizes performance. Our main claims rest on the controlled re-runs (top block).

The Schema-Guided Agent achieves the highest EM and F1 among controlled methods on every dataset, outperforming AtomR by +2.6 EM on BlendQA, +19.8 on HybridQA, +5.7 on TAT-QA, and +10.5 on ComplexTR. On HybridQA and ComplexTR, where ProbTree is the strongest controlled baseline, the margins are +2.8 and +4.5 EM.
Gains are largest where structured evidence must be aligned with textual evidence: \textsc{TAT-QA} (+55.8 EM over RAG) and \textsc{HybridQA} (+51.3 EM). On ComplexTR, HippoRAG~2 is ahead on LJ (81.5 vs.\ 76.3), while we lead on F1 (82.2 vs.\ 78.2) and BLEU (82.0 vs.\ 72.7). Error analysis suggests rare and temporally reified types are harder to realize in KG construction, reducing coverage for some time-scoped queries (Appendix~\ref{app:failure-analysis}).

\subsection{Ablation Studies}
\label{sec:ablations}

\paragraph{Component ablations (Table~\ref{tab:qa_all_metrics}, bottom).}
In ``w/o schema guidance,'' schema-conditioned routing is disabled; on \textsc{BlendQA} this scores 20.0 EM -- nearly identical to ReAct (20.4 EM, Table~\ref{tab:agent_diagnostics}), confirming that without schema conditioning the additional infrastructure provides no routing advantage ($-$9.3 EM on \textsc{BlendQA}, $-$2.1 to $-$7.3 on other datasets).
\textbf{KG ingestion} contributes most on multi-hop tasks ($-$8.0 EM on \textsc{BlendQA}, $-$5.7 on \textsc{ComplexTR}) and least on the table-centric ones ($-$0.5 on \textsc{HybridQA}, $-$2.6 on \textsc{TAT-QA}), where schema-guided structured lookup carries most of the load. Comparing the full system and the w/o-KG ablation question by question, the graph turns 10.2\% of BlendQA answers from wrong to right and 2.2\% from right to wrong, mostly on cross-source multi-hop questions (Appendix~\ref{app:kg-flips}).
\textbf{Structural intelligence}: removing it drops EM by 2.1--6.8 across datasets.
\textbf{Schema extension}: removing it drops EM by 2.8 on \textsc{HybridQA} and 1.5 on \textsc{TAT-QA}, so most of the gain does not rely on query-time extension.
Component importance varies by dataset: schema-conditioned routing is most visible on \textsc{BlendQA} and \textsc{ComplexTR}, while \textsc{HybridQA} and \textsc{TAT-QA} benefit substantially from schema-compatible structured evidence access even without routing guidance.

\begin{table}[t]
\centering
\small
\setlength{\tabcolsep}{3pt}
\begin{tabular}{lccccc}
\toprule
\textbf{Agent} & \textbf{EM} & \textbf{F1} & \textbf{BLEU} & \textbf{LJ} & \textbf{Latency} \\
\midrule
\multicolumn{6}{l}{\textit{GPT-4.1}} \\
RAG (E5) & 19.1 & 33.6 & 26.1 & 51.0 & 0.3s \\
RAG (E5+rerank) & 20.2 & 34.8 & 32.4 & 59.7 & 0.4s \\
ReActAgent & 20.4 & 41.8 & 42.6 & 63.5 & 1.5s \\
PlanActAgent & 20.1 & 40.2 & 39.8 & 62.6 & 1.7s \\
SchemaAgent (ours) & \textbf{29.3} & \textbf{48.9} & \textbf{54.1} & \textbf{70.0} & 3.2s \\
\midrule
\multicolumn{6}{l}{\textit{Claude Haiku 4.5}} \\
RAG & 20.1 & 33.4 & 27.2 & 55.1 & -- \\
ReAct & 24.7 & 45.0 & 43.2 & 66.2 & -- \\
SchemaAgent (ours) & \textbf{28.1} & \textbf{46.1} & \textbf{45.7} & \textbf{69.4} & -- \\
\midrule
\multicolumn{6}{l}{\textit{Llama 3.3 70B}} \\
RAG & 13.9 & 23.0 & 21.9 & 30.4 & -- \\
ReAct & 17.7 & 28.0 & 36.3 & 41.4 & -- \\
SchemaAgent (ours) & \textbf{24.0} & \textbf{36.7} & \textbf{44.5} & \textbf{67.1} & -- \\
\bottomrule
\end{tabular}
\caption{Agent architectures and LLM backbones on BlendQA (445 multi-source questions): retrieval baselines, generic agents (ReAct, PlanAct), and our schema-guided agent. Latency is reported for GPT-4.1 only. LJ uses the Claude Haiku~4.5 judge for all backbones. }
\label{tab:agent_diagnostics}
\label{tab:model_results_main}
\vspace{-2mm}
\end{table}

\paragraph{Agent architecture comparison (Table~\ref{tab:agent_diagnostics}).}
Generic tool-use agents (ReAct, PlanAct) barely improve over RAG ($+$1.0--1.3 EM) despite 5--6$\times$ latency.
The Schema-Guided Agent achieves +10.2 EM and +15.3 F1 over RAG, confirming gains stem from \emph{schema-conditioned routing} rather than agentic iteration alone.
This is a latency--accuracy trade-off: the schema-guided agent takes 3.2\,s per query versus 0.3\,s for RAG.

\subsection{Cross-Model Evaluation}
\label{sec:model-agnostic-main}

On BlendQA with three backbone families (Table~\ref{tab:model_results_main}), schema guidance improves over RAG and ReAct in all three, by +8.0 EM over RAG (Claude Haiku~4.5), +10.1 (Llama~3.3~70B), +10.2 (GPT-4.1), so gains are not specific to GPT-4.1. Appendix~\ref{app:crossmodel-tatqa} shows the same pattern on TAT-QA. 
with Claude Haiku~4.5 and Llama~3.1~70B.

\subsection{Schema/KG Quality and Online Serving}
\label{sec:schema-kg-eval}
\label{sec:online-perf}

Across nine intrinsic metrics (Appendix~\ref{app:schema-kg-metrics}), link validity and provenance completeness are 100\% on all datasets and type utilization is 50--89\%. The main bottleneck is schema realization: some valid types, especially rare or temporally reified ones, are not populated, which limits graph coverage but not answerability, since the agent also uses structured lookup and vector evidence (failure modes in Table~\ref{tab:failure-modes}). Schema generation completes in 0.5--2.4\,min and ingestion in 0.33--3.3\,s per document (Appendix~\ref{app:schema-kg-metrics}).
In online serving (HybridQA; Table~\ref{tab:online_metrics}), KG traversal (4.6\,ms) and schema lookup (0.9\,ms) add little latency; vector retrieval (584\,ms) and LLM synthesis (707\,ms) dominate the 1.85\,s base query, and extension, when triggered (Section~\ref{sec:schema-ext}), adds ${\sim}$3.4\,s.

\section{Conclusion}
We presented a system that treats an automatically induced schema as a shared executable contract for both KG ingestion and query-time retrieval. In controlled comparisons across four benchmarks, ablations confirm that schema guidance, structural intelligence, and schema-guided construction each contribute to QA accuracy. Cross-model evaluation on BlendQA shows consistent gains across three LLM families.

Our results suggest three principles for heterogeneous QA: (1)~closed-world schema induction makes LLM-discovered structure executable; (2)~statistical structural analysis should precede semantic extraction; and (3)~retrieval agents benefit when the construction schema also conditions routing -- on BlendQA, this substantially outperforms generic ReAct/PlanAct tool use, and the full system leads controlled baselines on all four benchmarks.

\section*{Limitations}

\textbf{Scope of cross-model evaluation.}
We evaluate three LLM families on BlendQA and TAT-QA and find consistent gains (Section~\ref{sec:model-agnostic-main}, Appendix~\ref{app:crossmodel-tatqa}). Extending this to HybridQA and ComplexTR and to smaller open-weight models would further strengthen generalizability claims.

\noindent \textbf{Schema realization vs.\ coverage.}
The closed-world field catalog prevents hallucinated field references but does not prevent spurious relationships among valid fields. Schema type utilization ranges from 50\% to 89\% across datasets. Not all schema-defined types are reliably populated during KG construction, particularly rare or domain-specific ones; however, the agent mitigates this through fallback to structured lookup and vector retrieval, so unrealized types reduce graph-based coverage without eliminating answerability (Appendix~\ref{app:schema-kg-metrics}).

\noindent \textbf{Extension latency.}
Query-time schema extension adds ${\sim}$3.4\,s when triggered (98.4\% utilization rate on HybridQA), which may be prohibitive for real-time applications with strict latency requirements.

\noindent \textbf{Evaluation scope.}
We evaluate on four publicly available benchmarks with diverse characteristics (Table~\ref{tab:datasets}). Generalization to production enterprise data with noisier sources, larger scale, and stricter governance requirements has not been validated.

\bibliography{bibliography}

\appendix
\input{supplementary}

\end{document}

%% file: supplementary.tex

\clearpage
\section{Schema Generation Pipeline: Technical Details}
\label{app:schema}

This appendix provides implementation details and pseudocode for the schema generation pipeline described in Section~\ref{sec:schema-gen}.
Given a collection of sources $\mathcal{S}=\{S_1,\dots,S_m\}$, the system induces per-source schemas $\{\Sigma_{S_i}\}$ and unifies them into a global schema $\Sigma$.

\subsection{Notation and Definitions}
\label{app:schema:notation}

We adopt the following notation throughout this appendix:
\begin{itemize}
  \item For a structured source $S$, let $\{r_1,\dots,r_n\}$ denote sampled records (or files, for corpus sources).
  \item Let $\mathcal{F}_S$ be the \emph{closed-world field catalog}: the set of all observable fields in $S$ (columns for tabular sources; JSON paths for semi-structured sources).
  \item Let $r^\star$ denote the \emph{representative sample} selected for deep discovery.
  \item Let $\mathcal{D} = \{d_1, \dots, d_k\}$ denote the \emph{diverse samples} used for validation.
  \item For a field $f$, let $V_f$ denote the multiset of observed values across sampled records.
\end{itemize}

\paragraph{Core artifacts.}
\textbf{Schema $\Sigma$ (YAML).} A versioned artifact containing the field catalog $\mathcal{F}$ with per-field statistics, entity/relationship definitions grounded in $\mathcal{F}$, structural intelligence $\mathcal{I}$, and cross-source link specifications.
\textbf{KG and evidence store $G$.} A Neo4j property graph with typed domain entities and \texttt{SourceRecord} nodes; all elements connect to source records via \texttt{DERIVED\_FROM} edges ($\pi$). Unstructured evidence is chunked and indexed for dense retrieval.

\subsection{Pipeline Overview}
\label{app:schema:overview}

The schema generation pipeline processes each source $S \in \mathcal{S}$ through seven stages:

\begin{enumerate}
  \item \textbf{Source discovery:} Scan the input directory, detect file formats, and group files into logical sources.
  \item \textbf{Profiling:} For each source, enumerate observable fields and compute statistics (types, null rates, cardinalities).
  \item \textbf{Field catalog construction:} Build the closed-world catalog $\mathcal{F}_S$ with stable field identifiers.
  \item \textbf{Sampling:} Select representative sample $r^\star$ and diverse set $\mathcal{D}$.
  \item \textbf{Two-pass LLM discovery:} Extract entities and relationships via two targeted LLM calls.
  \item \textbf{Structural inference:} Detect identity keys, foreign keys, and hierarchy using statistical methods (no LLM).
  \item \textbf{Cross-source linking and unification:} Discover inter-source relationships and merge per-source schemas into $\Sigma$.
\end{enumerate}

The remainder of this section details each stage.

\subsection{Source Discovery and Corpus Detection}
\label{app:schema:discovery}

The pipeline begins by scanning the input directory to identify data sources. We support JSON, JSONL, CSV, and Excel files. A key challenge is handling \emph{corpus sources} -- directories containing thousands of individual text files with homogeneous structure.

\paragraph{Corpus detection heuristic.}
A directory containing at least 50 files is flagged as a corpus candidate. We then sample $k{=}5$ files and compute a structural signature for each (top-level JSON key set or CSV column names); if at least 80\% of the sampled files share the same signature, the directory is classified as a corpus source. Rather than treating each file as a separate source (which would be computationally prohibitive), we treat the entire directory as a single logical source with file-level sampling. This design reflects our observation that corpus directories typically contain homogeneous documents sharing a common schema.

\paragraph{Pseudocode.}
Algorithm~\ref{alg:discover} outlines the source discovery procedure.

\begin{algorithm}[t]
\caption{Source Discovery}
\label{alg:discover}
\begin{algorithmic}[1]
\Procedure{DiscoverSources}{$\mathit{root}$}
  \State $\mathcal{S} \gets \emptyset$
  \For{each path $p$ in \Call{RecursiveList}{$\mathit{root}$}}
    \If{\Call{IsDirectory}{$p$} \textbf{and} \Call{CountFiles}{$p$} $\geq 50$
          \textbf{and} \Call{StructSim}{$p$} $\geq 0.8$}
      \State $\mathcal{S} \gets \mathcal{S} \cup \{\textsc{CorpusSource}(p)\}$
    \ElsIf{\Call{IsSupportedFile}{$p$}}
      \State $\mathcal{S} \gets \mathcal{S} \cup \{\textsc{FileSource}(p)\}$
    \EndIf
  \EndFor
  \State \Return \Call{GroupCompatibleSources}{$\mathcal{S}$}
\EndProcedure
\end{algorithmic}
\end{algorithm}

\subsection{Profiling and Sampling}
\label{app:schema:profiling}

After discovering sources, we profile each source to extract field metadata and select samples for LLM analysis. The goal is to minimize LLM cost while maximizing schema coverage.

\subsubsection{Field Profiling}
For each source $S$, profiling extracts:
\begin{itemize}
  \item \textbf{Field paths:} All observable JSON paths or column names.
  \item \textbf{Data types:} Inferred from value patterns (string, number, boolean, array, object).
  \item \textbf{Null rates:} Fraction of records where each field is absent or null.
  \item \textbf{Example values:} A small sample of non-null values for each field.
  \item \textbf{Cardinality:} Number of unique values observed.
\end{itemize}

These statistics inform both the sampling strategy and subsequent structural analysis.

\subsubsection{Representative Sample Selection}
We select a single representative sample $r^\star$ for deep LLM analysis.  Each candidate is scored by three heuristics:

\begin{equation}
\label{eq:rep-score}
\begin{split}
\mathit{score}(r) \;=\; & \underbrace{10 \cdot |\{f : r[f] \neq \texttt{null}\}|}_{\text{field completeness}} \\
                        & \;+\; \underbrace{b_{\text{rich}}(r)}_{\text{content richness}} \;+\; \underbrace{b_{\text{pos}}(r)}_{\text{positional bonus}}
\end{split}
\end{equation}

\noindent where $b_{\text{rich}} = +50$ for text content $>$1\,K chars, $-30$ for $<$100 chars, and $b_{\text{pos}} = +10$ when the record lies in the middle 60\% of the file.  The candidate with the highest score is selected as $r^\star$.

\subsubsection{Diverse Sample Selection}
We select diverse samples via positional sampling (beginning, middle, end) and quality-aware coverage-adaptive selection.  Samples are packed into the remaining LLM token budget ($\sim$8K tokens) after accounting for the field catalog and prompt instructions, so the effective count adapts to record size.

\subsection{Closed-World Field Catalog}
\label{app:schema:catalog}

A critical component of our approach is the \emph{closed-world field catalog} $\mathcal{F}_S$. Before invoking the LLM, we enumerate all observable fields and assign each a stable identifier. The LLM is then constrained to reference only fields from this catalog, preventing hallucinated structure.

\subsubsection{Catalog Structure and Validation}
Each entry in $\mathcal{F}_S$ stores: a stable hash-based ID, the source name, the field path (or column name), the inferred data type, null rate, example values, and optional semantic tags.
The LLM is constrained to reference only catalog field IDs; we validate all references:

\begin{equation}
\label{eq:field-validity}
\mathit{FieldValidity} = \frac{|\{\text{referenced IDs}\} \cap \mathcal{F}_S|}{|\{\text{referenced IDs}\}|}
\end{equation}

Schemas with $\mathit{FieldValidity} < 1.0$ are flagged for repair. The closed-world constraint prevents hallucinated field references.

\subsection{Two-Pass LLM Discovery}
\label{app:schema:llm}

We extract entities and relationships using exactly two LLM calls per source. This design balances extraction quality against cost: a single call often misses edge cases, while many calls become prohibitively expensive.

\textbf{Call 1 (Deep Discovery).} Analyzes a token-budget--packed set of representative samples together with the full field catalog.  The LLM extracts entity types, attributes (mapped to catalog field IDs), and relationships.

\textbf{Call 2 (Validation).} Reviews diverse samples $\mathcal{D}$ (selected by positional sampling) to identify entity types or relationships not visible in the first call.

Outputs from both calls are merged (Section~\ref{app:schema:llm:merge}), yielding broader coverage than a single-pass call -- particularly for rare entity types that appear only in a minority of records.

\subsubsection{Output Merging and Linting}
\label{app:schema:llm:merge}
New entities from Call~2 are appended, new attributes extend existing entity definitions, and relationships are deduplicated by (source, target, type) tuple. A schema linter (a rule-based checker, as in software linting) then validates:
\begin{itemize}
  \item All field references exist in $\mathcal{F}_S$.
  \item All relationship endpoints reference defined entity types.
  \item Entity and relationship names follow naming conventions (PascalCase for entities, UPPER\_SNAKE for relationships).
\end{itemize}

\subsection{Structural Intelligence}
\label{app:schema:struct}

Beyond LLM-discovered semantics, we infer structural patterns using statistical analysis of sampled values. This non-LLM analysis detects identity keys, foreign keys, and source hierarchy -- patterns that are difficult for LLMs to detect reliably but straightforward to compute from data statistics.

\subsubsection{Identity Key Detection}
Identity keys are fields (or field combinations) that uniquely identify entity instances, enabling deterministic deduplication during ingestion.

A field $f$ is accepted as an identity key if it passes a two-stage gate.
First, a \emph{qualifying} check:
\begin{equation}
\label{eq:id-gate}
\begin{split}
\mathit{qualify}(f) = & \bigl[\mathit{uniq}(f) \geq 0.95\bigr] \\
                      & \;\lor\; \bigl[\mathit{uniq}(f) \geq 0.80 \;\land\; \mathit{isIdLike}(f)\bigr]
\end{split}
\end{equation}

\noindent where $\mathit{uniq}(f) = |V_f^{\text{unique}}| / |V_f|$ and $\mathit{isIdLike}$ matches patterns such as \texttt{id}, \texttt{key}, \texttt{uuid}, \texttt{ticker}, and \texttt{code}.
Fields with fewer than 5 non-null values, average length $>$500 characters, or cardinality $\leq 2$ are excluded.
For qualifying fields, a confidence score is assigned:
\begin{equation}
\label{eq:id-conf}
\mathit{conf}(f) = \min\!\bigl(0.95,\; 0.7 + 0.3 \cdot \mathit{uniq}(f)\bigr)
\end{equation}

\noindent The pipeline also detects \emph{composite keys} (pairs of fields whose combined uniqueness $\geq 0.95$), common in datasets keyed by (\texttt{ticker}, \texttt{quarter}).

\subsubsection{Foreign Key Detection}
Foreign keys connect entities across sources. For each pair of sources $(S_A, S_B)$, candidate columns are identified by exact or semantic name matching (e.g., \texttt{ticker} $\leftrightarrow$ \texttt{symbol}). For matched columns we compute value overlap:

\begin{equation}
\label{eq:fk-overlap}
\mathit{overlap}(f_A, f_B) = \frac{|V_{f_A} \cap V_{f_B}|}{\max(|V_{f_A}|, |V_{f_B}|)}
\end{equation}

\noindent Column pairs with non-zero overlap (or key-like names even without sampled overlap) are emitted as candidate foreign keys, with confidence $\min\bigl(0.95,\; 0.5 + 0.3 \cdot \mathit{overlap} + 0.15 \cdot \mathbf{1}_{\text{name}}\bigr)$, where $\mathbf{1}_{\text{name}}=1$ if the two columns have identical names and 0 otherwise.
Cardinality is inferred per shared key value: if the average record count per key in a source $\leq 1.2$ (and max $\leq 2$), that side is classified as ``one''; otherwise ``many''.  This yields cardinalities 1:1, 1:N, N:1, or N:M.

\subsubsection{JSON Path Detection}
For semi-structured sources, we detect nested arrays and objects that require special extraction handling. We recursively traverse sample structures and emit JSON path specifications:
\begin{itemize}
  \item Array at path \texttt{p}: emit \texttt{p[*]}
  \item Nested object with common array key: emit \texttt{p.items[*]}
\end{itemize}

These paths are included in the schema to guide the ingestion pipeline's extraction logic.

\subsection{Cross-Source Link Detection}
\label{app:schema:crosslink}

For multi-source datasets, we detect relationships between entities in different sources. Naive parallel indexing (comparing the $i$-th record from source $A$ with the $i$-th record from source $B$) fails for 1:N and N:M relationships. Instead, we use \emph{matched sampling}.

\subsubsection{Matched Sampling Algorithm}
The matched sampling procedure works as follows:
\begin{enumerate}
  \item \textbf{Discover link keys:} Identify candidate fields that connect sources (using the foreign key detection from Section~\ref{app:schema:struct}).
  \item \textbf{Group by key value:} For each source, group sampled records by their link key value.
  \item \textbf{Find common values:} Identify key values that appear in both sources.
  \item \textbf{Load matched sets:} For each common value, retrieve all records from both sources sharing that value.
  \item \textbf{Analyze matched pairs:} The LLM analyzes matched record sets to discover semantic relationships.
\end{enumerate}

Algorithm~\ref{alg:matched} provides pseudocode for this procedure.

\begin{algorithm}[t]
\caption{Matched Sampling for Cross-Source Links}
\label{alg:matched}
\begin{algorithmic}[1]
\Procedure{MatchedSampling}{$S_A, S_B, f_{\mathit{link}}$}
  \State $G_A \gets \Call{GroupBy}{\mathit{samples}_A, f_{\mathit{link}}}$
  \State $G_B \gets \Call{GroupBy}{\mathit{samples}_B, f_{\mathit{link}}}$
  \State $K_{\mathit{common}} \gets \Call{Keys}{G_A} \cap \Call{Keys}{G_B}$
  \State $\mathit{matched} \gets \emptyset$
  \For{$k \in \Call{Sample}{K_{\mathit{common}}, 5}$}
    \State $\mathit{matched} \gets \mathit{matched} \cup \{(G_A[k], G_B[k], k)\}$
  \EndFor
  \State \Return $\mathit{matched}$
\EndProcedure
\end{algorithmic}
\end{algorithm}

This approach correctly handles all cardinalities: for a given key value, we retrieve \emph{all} matching records from both sources, revealing the true relationship structure.

\subsubsection{Companion Source Detection}
As a preprocessing step, we detect \emph{companion sources} -- sources with related filenames (e.g., \texttt{queries.jsonl} and \texttt{queries\_metadata.jsonl}). These are linked via their implicit shared identifier before general cross-source detection.

\subsection{Schema Unification}
\label{app:schema:unify}

After processing all sources, we unify per-source schemas into a global schema $\Sigma$. Unification proceeds in three phases:

\paragraph{Entity merging.}
Entities appearing in multiple sources are matched by \emph{normalized name} (case-insensitive).  When two per-source schemas declare the same entity type (e.g., \texttt{WikidataEntity} in two files), they are merged: the first-encountered name becomes canonical, and attributes are unioned across sources.  Aliases (original, source-specific names) are retained for provenance.

\paragraph{Relationship deduplication.}
Relationships are deduplicated by (source\_entity, target\_entity, type) tuple. When duplicates exist, we retain the instance with higher confidence.

\paragraph{Provenance layer.}
We add a \texttt{SourceRecord} entity type to track provenance. Each ingested entity receives a \texttt{DERIVED\_FROM} relationship to its originating \texttt{SourceRecord}, enabling lineage queries.

\subsection{Ground-Truth Exclusion and Leakage Control}
\label{app:schema:gt}

For evaluation datasets, we must prevent ground-truth leakage: the schema should not expose answer fields that would trivialize the QA task. The pipeline accepts an exclusion configuration (a lightweight YAML manifest) specifying files and fields to hide.

\subsubsection{Exclusion Processing}
The \texttt{GTFilter} component applies exclusions at multiple stages:
\begin{enumerate}
  \item \textbf{Source discovery:} Excluded files are removed from the source list.
  \item \textbf{Sampling:} Excluded fields are stripped from all samples before LLM analysis.
  \item \textbf{Schema finalization:} Entities and attributes referencing excluded fields are removed. Relationships with orphaned endpoints (referencing removed entities) are also removed.
\end{enumerate}

\subsubsection{Per-Dataset Leakage Control}
Table~\ref{tab:leakage-control} details the fields and files excluded for each evaluation dataset to prevent ground-truth leakage.

\begin{table}[t]

\setlength{\tabcolsep}{2pt}

\centering
\footnotesize
\begin{tabularx}{\columnwidth}{@{}l>{\raggedright\arraybackslash}p{2.05cm}>{\raggedright\arraybackslash}X@{}}
\toprule
\textbf{Dataset} & \textbf{Type} & \textbf{Excluded Fields/Files} \\
\midrule
\textsc{BlendQA} & hybrid\_qa\_kb & \texttt{question}, \texttt{sub\_q1}, \texttt{sub\_q2}, \texttt{long\_answer}, \texttt{short\_answer}, \texttt{answers}. \\
\midrule
\textsc{HybridQA} & qa\_evaluation & \texttt{question}, \texttt{question\_id}, \texttt{question\_postag}, \texttt{answer-text}, \texttt{answer-node}. \\
\midrule
\textsc{TAT-QA} & hybrid\_qa\_kb & \texttt{questions} (entire array, containing \texttt{answer}, \texttt{derivation}, \texttt{scale}, \texttt{answer\_type}, \texttt{answer\_from}, \texttt{rel\_paragraphs}). Files: \texttt{*\_test*.json}, \texttt{*\_dev.json}. \\
\midrule
\textsc{ComplexTR} & qa\_evaluation & \texttt{question}, \texttt{answers}, \texttt{text\_answers}, \texttt{fact\_context}. \\
\bottomrule
\end{tabularx}
\caption{Leakage control: excluded fields and files per dataset}
\label{tab:leakage-control}
\end{table}

This design allows raw benchmark files to be processed without modifying
the underlying data, while using a lightweight exclusion manifest to hide
annotation fields from schema induction and ingestion.

\subsubsection{Leakage Audit: What Each Pipeline Stage Sees}
\label{app:schema:audit}

We provide a stage-by-stage audit of what information each component accesses, confirming that no ground-truth answers, derivation programs, scale annotations, or evaluation labels enter the schema, KG, or agent prompts.

\paragraph{Schema generation.}
The field catalog and LLM discovery calls see only KB fields. For TAT-QA, ingestion uses JSONPath selectors (\texttt{\$.table}, \texttt{\$.paragraphs[*]}) that are structurally disjoint from \texttt{\$.questions[*]} -- the GT array cannot be accessed through these paths regardless of field-name filtering. For HybridQA and ComplexTR, QA files are excluded at the file level before source discovery; only KB files (Wikipedia tables, corpus passages) enter the pipeline.

\paragraph{KG construction.}
Extraction, deduplication, and linking operate on schema-specified fields only. The schema may include field-level statistics and examples from the allowed evidence corpus, but it does not include gold answer fields, derivations, scale labels, answer types, answer provenance labels, or per-question supervision. KG nodes and vector chunks may contain values from allowed tables and paragraphs -- because those are the evidence sources all systems are expected to use -- but they do not contain gold answer annotations or per-question answer labels from the excluded ground-truth subtree (e.g., \texttt{questions[*]} in TAT-QA).

\paragraph{Query-time extension.}
Extension operates via two paths, neither of which accesses GT:
\begin{itemize}[nosep]
  \item \emph{Query-scoped extension} receives only the query string and the existing schema vocabulary. An LLM proposes candidate entity \emph{type definitions} (e.g., ``add type \texttt{Number}'') with confidence scores; it never sees raw data records, answer fields, or evaluation labels.
  \item \emph{Document-based extension} receives KB content (tables, paragraphs) after GT filtering via the centralized \texttt{GTFilter} module, which strips excluded fields before any LLM call. Extension proposes schema \emph{elements} (entity types, relationship types, attributes), not answer candidates.
\end{itemize}
\noindent In both modes, extension can only add schema \emph{definitions} (type names, relationship types, attribute lists); it cannot add KG nodes, entity values, or answer candidates. The output is a set of typed placeholders that guide downstream extraction, not retrieved evidence itself.

\paragraph{Extension utility metrics.}
We define three metrics for query-time extension (Table~\ref{app:online-serving-table}):
\textbf{ETR} (Extension Trigger Rate) is the fraction of queries where the sufficiency gate (Algorithm~\ref{alg:query-sufficiency}) detects unresolved entity types and triggers extension.
\textbf{EUR} (Extension Utilization Rate) is the fraction of triggered extensions for which at least one newly added schema type or relationship is used by a downstream retrieval step (measuring retrieval utilization, not answer-causal contribution).
\textbf{EWR} (Extension Waste Rate) $= 1 - \text{EUR}$: the fraction of triggered extensions that added types unused by retrieval.

\subsection{Complexity Analysis}
\label{app:schema:complexity}

Table~\ref{tab:complexity} summarizes the computational complexity of each pipeline stage.

\begin{table}[t]
\centering
\footnotesize
\setlength{\tabcolsep}{3pt}
\begin{tabularx}{\columnwidth}{@{}ll>{\raggedright\arraybackslash}X@{}}
\toprule
\textbf{Stage} & \textbf{Complexity} & \textbf{Notes} \\
\midrule
Source discovery & $O(F)$ & $F$ = total files \\
Profiling & $O(n \cdot k)$ & $n$ = samples, $k$ = fields \\
Sampling & $O(n \log n)$ & Sorting for median \\
LLM discovery & $O(m)$ & $m$ = sources; 2 calls each \\
Structural analysis & $O(m^2 \cdot k)$ & FK detection is pairwise \\
Cross-source links & $O(m^2 \cdot n)$ & Matched sampling \\
Unification & $O(E)$ & $E$ = total entities (name-based) \\
\bottomrule
\end{tabularx}
\caption{Complexity of schema generation stages}
\label{tab:complexity}
\end{table}

In practice, LLM calls dominate wall-clock time. With 2 calls per source and typical response times of 5--15 seconds, schema generation for a 10-source dataset completes in 2--5 minutes.

\subsubsection{Reproducibility Checklist}
\label{app:schema:reproducibility}

To reproduce schema generation:
\begin{enumerate}
  \item \textbf{Configuration:} Specify ground-truth exclusions for the target dataset via a YAML manifest.
  \item \textbf{Pipeline stages:} The schema generation script invokes the following modules in sequence:
  \begin{itemize}
    \item Multi-source orchestrator (coordinates per-source analysis)
    \item LLM-based dataset analyzer (two-pass entity/relationship discovery)
    \item Structural analyzer (identity keys, foreign keys, JSON paths)
    \item Semantic type enricher (canonical type mapping)
    \item Ground-truth filter (exclusion enforcement)
  \end{itemize}
  \item \textbf{Output:} Executable schema in YAML format.
  \item \textbf{Key parameters:} Sample size per source (default: 50), LLM model (GPT-4.1), field catalog enumeration (automatic), corpus file threshold (50), signature-match fraction (0.8, i.e., at least 4 of 5 sampled files)
\end{enumerate}

\section{KG Ingestion Pipeline: Technical Details}
\label{app:ingest}

\begin{figure*}[t]
    \centering
    \includegraphics[width=0.96\linewidth, trim={0.0cm 0.0cm 0.0cm 0.0cm},clip]{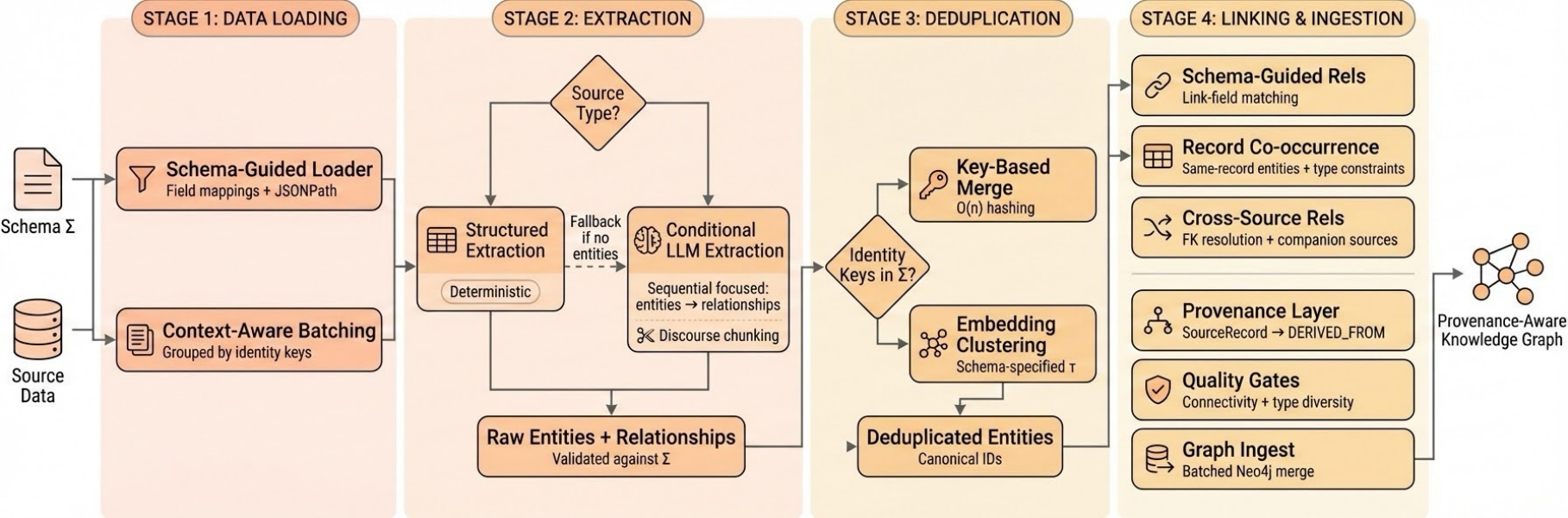}
    \vspace{-2mm}
    \caption{Schema-guided ingestion: extraction, deduplication, within/cross-source linking, and provenance edges.}
    \label{fig:kgingest}
    \vspace{-2mm}
\end{figure*}

This appendix provides implementation details for the schema-guided KG ingestion pipeline described in Section~\ref{sec:ingest-main}.
Given a schema $\Sigma$ and data sources $\mathcal{S}$, the pipeline produces a Neo4j knowledge graph with full provenance tracking.
Illustrative examples throughout this section use \textsc{BlendQA} (three heterogeneous sources sharing cross-source identifiers); the mechanisms generalize to all evaluation datasets.

\subsection{Pipeline Overview}
\label{app:ingest:overview}

Ingestion proceeds in five phases:
\begin{enumerate}
  \item \textbf{Source ordering:} Determine processing order from schema hierarchy (parents before children).
  \item \textbf{Entity extraction:} Apply multi-strategy extraction per source.
  \item \textbf{Deduplication:} Merge duplicate entities via identity keys and clustering.
  \item \textbf{Relationship extraction:} Build edges using three parallel strategies plus cross-source FK resolution.
  \item \textbf{Neo4j ingestion:} Write nodes and edges with provenance tracking.
\end{enumerate}

\subsection{Source Ordering}
\label{app:ingest:order}

When the schema defines hierarchical relationships between sources (e.g., parent entities referenced by child passages), we process parent sources first to ensure foreign key targets exist before children reference them.
Source hierarchy is inferred during schema generation (Section~\ref{app:schema:struct}): a source with fewer records linked via a 1:N key is classified as the parent.
For coordinated subsetting (e.g., during development), limits are propagated across sources sharing foreign key values to preserve referential integrity.

\subsection{Multi-Strategy Entity Extraction}
\label{app:ingest:extract}

Entity extraction uses a conditional decision tree rather than a fixed priority cascade.
Structured extraction is fast and free, while LLM extraction is slow and costly, so we invoke the LLM only when necessary.

\subsubsection{Decision Logic}

Algorithm~\ref{alg:extract-decision} formalizes the extraction decision.

\begin{algorithm}[t]
\caption{Entity Extraction Decision}
\label{alg:extract-decision}
\begin{algorithmic}[1]
\Procedure{ExtractEntities}{$\mathit{source}, \Sigma$}
  \State $E_{\mathit{struct}} \gets \Call{ExtractStructured}{\mathit{source}, \Sigma}$
  \State $\mathit{llm\_types} \gets \Sigma.\Call{GetLLMExtractableEntities}{\mathit{source}}$
  \State
  \State \textit{// Determine if LLM extraction needed}
  \If{$\Sigma.\Call{IsHybrid}{\mathit{source}}$}
    \State $\mathit{run\_llm} \gets \textbf{true}$ \Comment{Schema marks as hybrid}
  \ElsIf{$|E_{\mathit{struct}}| = 0$ \textbf{and} $|\mathit{llm\_types}| > 0$}
    \State $\mathit{run\_llm} \gets \textbf{true}$ \Comment{Fallback: no structured results}
  \Else
    \State $\mathit{run\_llm} \gets \textbf{false}$
  \EndIf
  \State
  \If{$\mathit{run\_llm}$}
    \State $E_{\mathit{llm}} \gets \Call{ExtractWithLLM}{\mathit{source}, \Sigma, \mathit{llm\_types}}$
  \Else
    \State $E_{\mathit{llm}} \gets \emptyset$
  \EndIf
  \State \Return $E_{\mathit{struct}} \cup E_{\mathit{llm}}$
\EndProcedure
\end{algorithmic}
\end{algorithm}

\subsubsection{Structured Extraction Methods}

Structured extraction supports three methods, automatically inferred if not specified:

\paragraph{Field mapping.}
Direct column-to-attribute mapping for structured fields.
Example from \textsc{BlendQA}:
\begin{lstlisting}[language=yaml]
WikidataEntity:
  extraction:
    method: field
    source_field: entity_id   # Wikidata QID (e.g., Q142)
  attributes: [entity_id, entity, concepts]
\end{lstlisting}

\paragraph{Nested field (JSONPath).}
Extraction from nested JSON structures.
The extractor iterates over JSONPath results and creates one entity per matching item:
\begin{lstlisting}[language=yaml]
WikidataTriple:
  extraction:
    method: nested_field
    json_path: "sub_source1[*]"
  attributes: [subject, predicate, object]
\end{lstlisting}

\paragraph{Row as entity.}
For sources where each row represents a complete entity (common after JSONPath expansion):
\begin{lstlisting}[language=yaml]
extraction:
  method: row_as_entity
attributes: [subject, predicate, object, source_qid]
\end{lstlisting}

\subsubsection{LLM Extraction with Batching and Chunking}

When LLM extraction is required, we optimize for throughput and cost:

\paragraph{Batching.}
Records are grouped into batches (default: 100 records) and processed in parallel across 8 workers, amortizing LLM call overhead while staying within API rate limits.

\paragraph{Semantic chunking.}
Long documents ($>$10K characters) are split into chunks of at most 8{,}000 characters with 500-character overlap.
The chunker prefers discourse-level boundaries in the following priority order:
\begin{enumerate}
  \item Section headers (HTML markers such as \texttt{<h1>}/\texttt{<h2>}, or markdown headings)
  \item Discourse boundaries (speaker turns, paragraph breaks)
  \item Sliding window (fallback)
\end{enumerate}

\paragraph{Co-extraction.}
During LLM entity extraction, we simultaneously extract relationships between co-occurring entities in each chunk.
This reduces total LLM calls by avoiding a separate relationship extraction pass.

\subsection{Two-Stage Deduplication}
\label{app:ingest:dedup}

Deduplication merges duplicate entities while preserving provenance.
We use a two-stage approach: deterministic identity key matching (fast, exact) followed by embedding-based clustering (slower, fuzzy).

\subsubsection{Stage 1: Identity Key Matching}

For entity types with identity keys in $\Sigma$, we group entities by key value and merge each group:

\begin{algorithm}[t]
\caption{Identity Key Deduplication}
\label{alg:dedup-id}
\begin{algorithmic}[1]
\Procedure{DeduplicateByKey}{$E, \mathit{key\_field}$}
  \State $G \gets \Call{GroupBy}{E, \lambda e: e.\mathit{attributes}[\mathit{key\_field}]}$
  \State $E' \gets \emptyset$
  \For{$(\mathit{key}, \mathit{group}) \in G$}
    \If{$|\mathit{group}| > 1$}
      \State $e_{\mathit{merged}} \gets \Call{MergeEntities}{\mathit{group}}$
    \Else
      \State $e_{\mathit{merged}} \gets \mathit{group}[0]$
    \EndIf
    \State $E' \gets E' \cup \{e_{\mathit{merged}}\}$
  \EndFor
  \State \Return $E'$
\EndProcedure
\end{algorithmic}
\end{algorithm}

Identity key matching groups entities by key value and merges each group in $O(n)$ time. When identity keys exist, this stage resolves most duplicates.

\subsubsection{Stage 2: Embedding-Based Clustering}

For entities without identity keys (or where key matching fails), we apply fuzzy matching:

\begin{enumerate}
  \item \textbf{Embed:} Convert each entity to text and compute sentence embeddings (\texttt{all-MiniLM-L6-v2}, 384-d).
  \item \textbf{Cluster:} Apply agglomerative clustering with distance threshold 0.15 (i.e., similarity $\geq 0.85$).
  \item \textbf{Merge:} For each cluster with $>$1 entity, merge into a canonical form preserving all source references.
\end{enumerate}

\subsubsection{ID Mapping}

Deduplication produces a mapping from temporary IDs to canonical IDs:
\[
\mathit{id\_map}: \mathit{temp\_id} \rightarrow \mathit{canonical\_id}
\]

This mapping is applied to all relationship endpoints before Neo4j ingestion, and duplicate relationships are removed.

\subsection{Three-Strategy Relationship Extraction}
\label{app:ingest:link}

Relationships are extracted using three parallel strategies whose results are merged.

\subsubsection{Strategy 1: Schema-Guided (Link Fields)}

When $\Sigma$ specifies \texttt{link\_fields} for a relationship type, we match entities deterministically:

\begin{lstlisting}[language=yaml]
# Example: link entities to their semantic types
relationships:
  - type: HAS_TYPE
    source_entity: WikidataEntity
    target_entity: SemanticType
    link_fields:
      - source_attr: concepts
        target_attr: type_name
\end{lstlisting}

\noindent For each source entity, the extractor finds target entities whose attribute value matches the specified source attribute, and creates typed edges.
These deterministic matches are assigned confidence 1.0.

\subsubsection{Strategy 2: Type-Based Patterns}

Structural relationships are inferred from entity type co-occurrence within the same source record.
When two entities of types for which $\Sigma$ defines a valid relationship are extracted from the same record, an edge is created.
For example, in \textsc{BlendQA}:

\begin{center}
\small
\begin{tabular}{lll}
\toprule
\textbf{Source Type} & \textbf{Target Type} & \textbf{Relationship} \\
\midrule
WikidataEntity & WikidataTriple & \texttt{HAS\_TRIPLE} \\
WikidataTriple & WikidataEntity & \texttt{OBJECT\_OF} \\
WebPassage & WikidataEntity & \texttt{MENTIONS} \\
$*$ & SourceRecord & \texttt{DERIVED\_FROM} \\
\bottomrule
\end{tabular}
\end{center}

\noindent These edges are assigned confidence 0.9--1.0, depending on the specificity of the type match.

\subsubsection{Strategy 3: LLM Extraction}

For relationships requiring semantic understanding, we invoke targeted LLM extraction using the original text context.
To control cost, LLM relationship extraction is capped at 50 calls per source, applied only to records containing 2+ co-occurring entities.

\begin{lstlisting}[language=yaml]
# Example: extract semantic relations from text
relationships:
  - type: DESCRIBED_IN
    source_entity: WikidataEntity
    target_entity: WebPassage
    extraction_method: llm
\end{lstlisting}

\noindent The LLM receives the source text, candidate entity mentions extracted from that text, and returns typed relationship instances with confidence scores.

\subsubsection{Cross-Source Foreign Key Resolution}

When sources share identifier columns (detected during schema generation, Eq.~\ref{eq:fk-overlap}), we resolve foreign keys by hash-based index joining:

\begin{algorithm}[t]
\caption{Foreign Key Resolution}
\label{alg:fk-resolve}
\begin{algorithmic}[1]
\Procedure{ResolveForeignKeys}{$E_A, E_B, \mathit{fk}$}
  \State $I_A \gets \Call{IndexBy}{E_A, \mathit{fk}.\mathit{source\_col}}$
  \State $I_B \gets \Call{IndexBy}{E_B, \mathit{fk}.\mathit{target\_col}}$
  \State $R \gets \emptyset$
  \For{$v \in \Call{Keys}{I_A} \cap \Call{Keys}{I_B}$}
    \For{$e_A \in I_A[v]$}
      \For{$e_B \in I_B[v]$}
        \State $R \gets R \cup \{(e_A, \mathit{fk}.\mathit{rel\_type}, e_B, \mathit{fk}.\mathit{conf})\}$
      \EndFor
    \EndFor
  \EndFor
  \State \Return $R$
\EndProcedure
\end{algorithmic}
\end{algorithm}

Cross-source foreign keys enable multi-hop queries spanning sources by linking entities via shared identifier values.
The FK resolver builds hash indexes over declared FK columns and materializes edges for matching values, with confidence inherited from the schema-discovery step.

\subsection{Provenance and SourceRecord Nodes}
\label{app:ingest:prov}

Every raw record is materialized as a \texttt{SourceRecord} node with full lineage tracking. Extracted entities receive \texttt{DERIVED\_FROM} edges to their originating records, enabling citation and grounding.

\subsubsection{Provenance Metrics and Methodology}
We measure provenance completeness as the fraction of extracted entities with valid \texttt{DERIVED\_FROM} edges. For grounding validation, we verify that:
\begin{enumerate}
  \item All entities have at least one \texttt{DERIVED\_FROM} edge to a \texttt{SourceRecord}.
  \item SourceRecord nodes preserve original field values for traceability (e.g., \texttt{entity\_id}, \texttt{record\_id}).
  \item Answer citations can be traced back to specific source records via graph traversal.
\end{enumerate}

The provenance layer enables answer grounding: when the QA agent returns an answer, it can cite the specific source records that supplied the supporting evidence via the \texttt{DERIVED\_FROM} relationship chain.

\subsection{Quality Validation and Dataset Preparation}
\label{app:ingest:quality}

\subsubsection{Quality Metrics}
Before serving, we compute lightweight quality metrics:

\begin{equation}
\label{eq:isolated}
\mathit{IsolatedRatio} = \frac{|\{v \in V : \deg(v) = 0\}|}{|V|}
\end{equation}

\begin{equation}
\label{eq:avgdeg}
\mathit{AvgDegree} = \frac{2|E|}{|V|}
\end{equation}

A graph is marked \emph{QA-ready} if $\mathit{IsolatedRatio} < 0.3$ and $\mathit{AvgDegree} \geq 2.0$.

\subsubsection{Schema and KG Evaluation Metrics}
\label{app:schema-kg-metrics}

\begin{table*}[t]
\centering
\small
\setlength{\tabcolsep}{3pt}
\scalebox{0.95}{%
\begin{tabular}{lc|cc|cc|cc|cc}
\toprule
\multirow{2}{*}{\textbf{Dataset}} &
\multicolumn{1}{c|}{\textbf{Block A}} &
\multicolumn{2}{c|}{\textbf{Block B}} &
\multicolumn{2}{c|}{\textbf{Block C}} &
\multicolumn{2}{c|}{\textbf{Block D}} &
\multicolumn{2}{c}{\textbf{Block E}} \\
& \textbf{XSrcUtil$\uparrow$}
& \textbf{SchUtil$\uparrow$} & \textbf{RelUse$\uparrow$}
& \textbf{LinkVal$\uparrow$} & \textbf{ProvCom$\uparrow$}
& \textbf{SchTypeAcc$\uparrow$} & \textbf{KGTypeAcc$\uparrow$}
& \textbf{SchTime} & \textbf{IngestTime} \\
\midrule
\textsc{BlendQA}               & 58\%  & 89\%  & 69\%  & 100\% & 100\% & 60\% & 58\% & 1.0 min & 3.3 sec/doc \\
\textsc{HybridQA}$^\dagger$   & 41\%  & 85\%  & 48\%  & 100\% & 100\% & 54\% & 49\%  & 1.4 min & 0.37 sec/doc \\
\textsc{TAT-QA}$^\dagger$     & 60\%  & 50\%  & 42\%  & 100\% & 100\% & 62\% & 42\% & 2.4 min & 0.33 sec/doc \\
\textsc{ComplexTR}             & N/A   & 75\%  & 43\%  & 100\% & 100\% & 60\% & 44\% & 0.5 min & 0.43 sec/doc \\
\bottomrule
\end{tabular}%
}
\caption{Unified schema+KG evaluation (5 blocks). ``N/A'' denotes not applicable. $^\dagger$Uses schema extension; Block~E times are one-time offline, amortized across all queries. IngestTime normalized per document; measured on subsets of up to 500 samples.}
\label{tab:schema_kg_metrics}
\end{table*}

\begin{table}[t]
\centering
\small
\begin{tabular}{lrrr}
\toprule
\textbf{Component} & \textbf{Avg (ms)} & \textbf{p50 (ms)} & \textbf{p90 (ms)}  \\
\midrule
KG Traversal & 4.6 & 4.2 & 6.3  \\
Schema Lookup & 0.9 & 0.8 & 1.5  \\
RAG/Vector & 584.1 & 550.7 & 1022.4 \\
LLM Synthesis & 706.6 & 673.2 & 923.8  \\
Other & 549.3 & -- & -- \\
\textit{Total (base query)} & \textit{1845.5} & \textit{1810.0} & \textit{2418.3}  \\
Extension latency & 3365 & 3270 & 4836 \\
\midrule
& \textbf{ETR} & \textbf{EUR} & \textbf{EWR}  \\
\midrule
Extension utility & 85.3\% & 98.4\% & 1.6\% \\
\bottomrule
\end{tabular}
\caption{Query latency on HybridQA (n=1000). ETR: Extension Trigger Rate, EUR: Utilization Rate, EWR: Waste Rate.}
\label{tab:online_metrics}
\label{app:online-serving-table}
\end{table}

We define nine evaluation metrics organized in five blocks (Table~\ref{tab:schema_kg_metrics}).

\paragraph{Block A: Schema planning quality.}
\textbf{CrossSourceUtil} (\textbf{XSrcUtil}) measures whether the source pairs $(S_i, S_j)$ a query needs are connected by at least one cross-source relationship in~$\Sigma$, reported as the fraction of evaluation queries for which this holds (required entity types are extracted per query without access to the schema, as in Block~D). High values indicate the schema provides join paths for multi-source queries.

\paragraph{Block B: Schema execution fidelity.}
\textbf{SchemaUtility} (\textbf{SchUtil}) is the fraction of entity types in~$\Sigma$ instantiated as at least one KG node. \textbf{RelationshipUsage} (\textbf{RelUse}) is the analogous fraction for relationship types. Together they quantify schema \emph{realization}: whether planned types were populated.

\paragraph{Block C: KG structural integrity.}
\textbf{LinkValidity} (\textbf{LinkVal}) is the fraction of edges whose source and target both exist -- necessary for traversability. \textbf{ProvenanceCompleteness} (\textbf{ProvCom}) is the fraction of domain entities connected via at least one \texttt{DERIVED\_FROM} edge to a source record, ensuring traceability.

\paragraph{Block D: End-to-end semantic support.}
\textbf{SchemaTypeAccuracy} (\textbf{SchTypeAcc}) is the fraction of entity types in evaluation queries that are semantically covered by the schema vocabulary. \textbf{KGTypeAccuracy} (\textbf{KGTypeAcc}) is the fraction of query-referenced types semantically covered by entity types present in the KG. We use a two-step protocol: (1)~an LLM extracts the abstract entity types needed to answer each query \emph{without access to the schema}; (2)~a separate LLM judge call evaluates whether each extracted type is semantically covered by any schema/KG type. This avoids both information leaking (extraction is schema-blind) and vocabulary-gap artifacts (the judge reasons about semantic equivalence, e.g., query-derived ``Revenue'' is covered by schema-derived ``FinancialMetric''). Both steps use GPT-4o-mini at temperature~$=0$. In preliminary checks, larger judges accepted broad schema types (e.g., Entity) as covering most query types, which saturates the metric; we therefore use the stricter judge, so reported values are conservative.

\paragraph{Block E: Automation efficiency.}
\textbf{SchTime} is wall-clock time for schema induction (field catalog, two-pass discovery, structural intelligence). \textbf{IngestTime} is per-document time from raw data to queryable Neo4j graph. Both measured on subsets of up to 500 samples; all Block~E times are \emph{one-time offline} and amortized across queries. Schema generation takes 0.5--2.4\,min and ingestion 0.33--3.3\,sec/doc (Table~\ref{tab:schema_kg_metrics}).

\subsubsection{Detailed Failure Analysis}
\label{app:failure-analysis}

Table~\ref{tab:failure-modes} summarizes the failure modes we observe, with evidence and an example for each. The first three emerge from the schema and KG evaluation metrics and are detailed below; the fourth comes from inspecting the KG and the last from agent traces.

\begin{table*}[t]
\centering
\small
\setlength{\tabcolsep}{4pt}
\begin{tabularx}{\linewidth}{@{}p{3.2cm}p{4.2cm}X@{}}
\toprule
\textbf{Failure mode} & \textbf{Evidence} & \textbf{Example} \\
\midrule
Relationship vocabulary redundancy & RelUse 42\% (\textsc{TAT-QA}) to 69\% (\textsc{BlendQA}) & \texttt{AT\_COMPANY} is never populated because extraction canonicalizes it into \texttt{MEMBERSHIP\_OR\_EMPLOYMENT\_AT\_ORGANIZATION}. \\
Uninstantiated schema types (over-specification) & SchUtil 50\% on \textsc{TAT-QA} & \texttt{ActuarialAssumption} is a valid type but too rare in \textsc{TAT-QA} to be extracted. \\
Entity resolution and type loss & KGTypeAcc ${<}$ SchTypeAcc on all datasets (\textsc{ComplexTR} 44\% vs.\ 60\%; \textsc{TAT-QA} 42\% vs.\ 62\%) & Without identity keys, embedding-based clustering discards borderline entities. \\
Spurious relationship among valid fields & 4 of 5 \texttt{HAS\_VALUE\_IN\_PERIOD} instantiations on \textsc{TAT-QA} attach year fields to non-metric types & \texttt{HAS\_VALUE\_IN\_PERIOD}(\texttt{ProductCategory}, \texttt{TimePeriod}) uses attested product-row and year-header fields, but the period-scoped value belongs to \texttt{FinancialMetric}. \\
Incorrect routing & Observed in agent traces; there is no routing ground truth from which to compute a rate & ``What was the nickname of the \ldots 1932 \ldots gold medal winner?'' is classified as a person lookup (\texttt{is\_multihop=false}) and returns Rudolf Svensson from the table instead of ``Starke Rudolf'' from the linked biography. \\
\bottomrule
\end{tabularx}
\caption{Observed failure modes. Schema/KG metrics are from Table~\ref{tab:schema_kg_metrics}.}
\label{tab:failure-modes}
\end{table*}

\paragraph{Relationship vocabulary redundancy.}
RelUse ranges from 42\% (\textsc{TAT-QA}) to 69\% (\textsc{BlendQA}).
Unused types fall into: (i)~\emph{semantic alternatives} canonicalized into a dominant pattern (e.g., \texttt{AT\_COMPANY} subsumed by \texttt{MEMBERSHIP\_OR\_EMPLOYMENT\_AT\_ORGANIZATION}), and (ii)~\emph{cross-source structural types} requiring FK signals absent in the data.
A pruning pass could yield RelUse closer to 100\%, but we retain unused types for extension and future ingestion.

\paragraph{Schema over-specification.}
Up to half of schema-defined entity types go uninstantiated (\textsc{TAT-QA} SchUtil\,$=$\,50\%): the two-pass LLM discovery proposes all plausible types, but extraction only populates those with clear structural signals or sufficient frequency.
Domain-specific types in specialized corpora (e.g., \texttt{ActuarialAssumption} in \textsc{TAT-QA}) appear too rarely for reliable extraction, leaving valid definitions unpopulated.

\paragraph{Entity resolution and type loss.}
When identity keys cannot be discovered, deduplication falls back to embedding-based clustering (cosine similarity ${\geq}0.85$).
KGTypeAcc falls below SchTypeAcc on every dataset, indicating schema-defined types lost during extraction and deduplication.
The gap is largest on \textsc{ComplexTR} (44\% vs.\ 60\%) and \textsc{TAT-QA} (42\% vs.\ 62\%), where fuzzy matching discards borderline entities, and smallest on \textsc{BlendQA} (58\% vs.\ 60\%), where heterogeneous sources provide richer extraction context.

\subsubsection{Dataset Preparation Methodology}
\paragraph{Table formatting.} Structured tables (CSV, Excel) are processed row-by-row. Nested JSON arrays within table cells are expanded using JSONPath specifications from the schema (e.g., \texttt{participants[*].name}).

\paragraph{Text chunking.} Long documents are split into semantic chunks with the following parameters:
\begin{itemize}
  \item \textbf{Chunking threshold:} 10{,}000 characters (shorter documents are processed as-is).
  \item \textbf{Max chunk size:} 8{,}000 characters (fits within LLM context windows while preserving semantic coherence).
  \item \textbf{Overlap:} 500 characters to preserve context across chunk boundaries.
  \item \textbf{Field truncation:} 12{,}000 characters per field (prevents extremely long values from overwhelming LLM prompts).
  \item \textbf{Chunking strategy:} Section headers $\rightarrow$ discourse boundaries (speaker turns, paragraphs) $\rightarrow$ sliding window (fallback).
\end{itemize}

\noindent The 10K threshold avoids unnecessary chunking for short documents while keeping longer inputs within LLM context limits.
The 8K chunk size ensures chunks fit within typical context windows (16K--32K tokens) with room for prompts and system messages.

\paragraph{Ingestion thresholds.}
\label{app:ingest:thresholds}
Key thresholds governing schema generation, extraction, deduplication, and linking, with the rationale for each. These values are design choices and were not tuned on the benchmarks.
\begin{itemize}
  \item \textbf{Corpus detection:} ${\geq}50$ files, with ${\geq}80\%$ of $k{=}5$ sampled files sharing one structural signature (Section~\ref{app:schema:discovery}). The 0.8 cutoff requires 4 of 5 sampled files to match, tolerating one auxiliary or malformed file. On our benchmarks, source directories hold either at most 5 files or over 15K (\textsc{HybridQA}'s per-table files, which share one signature in every sample), so the 50-file minimum separates single-file sources from per-record corpora with a wide margin on both sides.
  \item \textbf{Identity keys:} uniqueness ${\geq}0.95$, or ${\geq}0.80$ for ID-like field names (Eq.~\ref{eq:id-gate}). The 0.95 cutoff tolerates a small fraction of duplicates from sampling and dirty records; an identifier-like field name serves as prior evidence, so 0.80 suffices.
  \item \textbf{Entity clustering:} Embedding similarity $\geq 0.85$ (distance $\leq 0.15$) for fuzzy deduplication. Balances precision (avoiding false merges) against recall (catching surface-form variants).
  \item \textbf{Link field matching:} Deterministic, confidence 1.0, when the schema specifies explicit link fields.
  \item \textbf{Foreign key overlap:} Column pairs with overlap $>0$ (Eq.~\ref{eq:fk-overlap}) are emitted as candidate FKs during schema generation; confidence is $\min\bigl(0.95,\; 0.5 + 0.3\cdot\mathit{overlap} + 0.15\cdot\mathbf{1}_{\text{name}}\bigr)$. This score ranks and deduplicates candidates; it is not an acceptance threshold. Pairs with identical names rank above pairs whose names differ, and the highest possible score, 0.95, stays below schema-declared links (confidence 1.0).
  At ingestion time, declared FKs are resolved via hash-index joining.
  \item \textbf{Chunking:} 10K-character threshold and 8K-character chunks with 500-character overlap (rationale above, under \emph{Text chunking}).
  \item \textbf{LLM relationship cap:} Maximum 50 LLM calls per source for semantic relationship extraction, preventing excessive cost on sources with many entity co-occurrences.
\end{itemize}

\paragraph{Validation example (\textsc{BlendQA}).}
With 28 entity types and ${\sim}$5.2K nodes across three sources, the graph achieves high connectivity via shared identifier values (connectivity $>$99\%, average confidence 0.98; see Table~\ref{tab:ingest-perf} for per-dataset statistics).

\subsection{Neo4j Ingestion}
\label{app:ingest:neo4j}

Entities and relationships are written to Neo4j using batched operations.

\subsubsection{Index Creation}

For each entity type in $\Sigma$, we create Neo4j indexes on the node label and its identity key (if declared) plus any frequently queried attributes:
\begin{lstlisting}[basicstyle=\ttfamily\footnotesize]
CREATE INDEX IF NOT EXISTS FOR (n:<EntityType>) ON (n.id)
CREATE INDEX IF NOT EXISTS FOR (n:<EntityType>) ON (n.<identity_key>)
\end{lstlisting}

\subsubsection{Batched MERGE}

Entities are upserted via \texttt{MERGE} to ensure idempotency:
\begin{lstlisting}[basicstyle=\ttfamily\footnotesize]
MERGE (n:<EntityType> {id: $id})
ON CREATE SET n = $properties
ON MATCH SET n += $properties
\end{lstlisting}

\noindent We process entities in batches of 1{,}000 to balance throughput against memory.
Relationships are written similarly, with deduplication-mapped endpoint IDs.

\subsection{Performance Characteristics}
\label{app:ingest:perf}

Table~\ref{tab:ingest-perf} summarizes ingestion metrics for datasets where global KG ingestion statistics are collected.

\begin{table}[t]
\centering
\small
\setlength{\tabcolsep}{4pt}
\begin{tabular}{lccc}
\toprule
\textbf{Dataset} & \textbf{DetRatio} & \textbf{ConfAvg} & \textbf{DomRatio} \\
\midrule
\textsc{BlendQA}   & 79\% & 0.98 & 83\% \\
\textsc{TAT-QA}    & 75\% & 0.97 & 61\% \\
\textsc{ComplexTR}  & 51\% & 0.95 & 58\% \\
\bottomrule
\end{tabular}
\caption{KG ingestion metrics across evaluation datasets. DetRatio: fraction of relationships extracted via deterministic (non-LLM) methods. ConfAvg: mean relationship confidence. DomRatio: fraction of nodes that are domain entities (excluding \texttt{SourceRecord} nodes).}
\label{tab:ingest-perf}
\end{table}

\subsubsection{Reproducibility Checklist}
\label{app:ingest:reproducibility}

To reproduce KG ingestion:
\begin{enumerate}
  \item \textbf{Prerequisites:} Generated executable schema (from Section~\ref{app:schema}).
  \item \textbf{Pipeline stages:} The ingestion driver invokes the following modules:
  \begin{itemize}
    \item Pipeline orchestrator (batching, parallelism, progress tracking)
    \item Multi-strategy entity extractor (structured + conditional LLM extraction)
    \item Two-stage entity deduplicator (identity-key + embedding-based)
    \item Three-strategy relationship extractor (FK links, type patterns, LLM)
    \item Cross-source FK resolver
    \item Neo4j graph writer
  \end{itemize}
  \item \textbf{Configuration:} Neo4j connection via environment variables or a local settings file.
  \item \textbf{Key parameters:} Batch size (default: 100), max workers (default: 8), chunk size (8,000 chars), overlap (500 chars), similarity threshold (0.85), embedding model (\texttt{all-MiniLM-L6-v2}).
  \item \textbf{Output:} Neo4j graph with provenance edges (\texttt{DERIVED\_FROM}).
\end{enumerate}

\section{Schema Extension: Technical Details}
\label{app:extend}

This appendix provides implementation details for the schema extension system described in Section~\ref{sec:schema-ext}.

\subsection{Architecture Overview}
\label{app:extend:arch}

The extension system comprises two complementary components with different analysis depths:

\begin{enumerate}
  \item \textbf{SchemaExtender} (persistent): Permanently extends the schema when new documents arrive. Runs the full analysis pipeline -- two-pass LLM discovery, structural analysis (identity keys, foreign keys, JSON paths), and semantic type enrichment -- identical to initial schema generation (Section~\ref{app:schema}).
  \item \textbf{QueryDrivenSchemaExtender} (query-scoped): Extends the schema in-memory for a specific query with automatic revert. Uses a single LLM call to analyze query requirements and propose candidate entity types with confidence scores, trading analysis depth for speed (seconds rather than minutes).
\end{enumerate}

\subsection{Sufficiency Analysis}
\label{app:extend:sufficiency}

Before invoking expensive deep analysis, we perform a fast sufficiency check to determine whether extension is needed.

\subsubsection{Query-Based Sufficiency}

For user queries, we extract entity mentions and compare against the current schema:

\begin{algorithm}[t]
\caption{Query Sufficiency Analysis}
\label{alg:query-sufficiency}
\begin{algorithmic}[1]
\Procedure{AnalyzeQueryRequirements}{$Q, \Sigma$}
  \State $E_{\mathit{mentioned}} \gets \Call{ExtractEntityMentions}{Q}$
  \State $E_{\mathit{schema}} \gets \Call{GetEntityTypes}{\Sigma}$
  \State $\mathit{missing} \gets \emptyset$
  \For{$e \in E_{\mathit{mentioned}}$}
    \State $e_{\mathit{norm}} \gets \Call{Normalize}{e}$ \Comment{``sovereign state'' $\rightarrow$ ``Country''}
    \If{$e_{\mathit{norm}} \notin E_{\mathit{schema}}$}
      \State $\mathit{missing} \gets \mathit{missing} \cup \{e\}$
    \EndIf
  \EndFor
  \State \Return $(|\mathit{missing}| > 0, \mathit{missing})$
\EndProcedure
\end{algorithmic}
\end{algorithm}

Query sufficiency analysis extracts entity mentions from queries and checks against schema vocabulary. If all mentions map to existing entities, extension is skipped; otherwise, deep extension is triggered.

\subsubsection{Document-Based Sufficiency (Quick Analyze)}

For new documents, we offer a fast compatibility check sampling 5 records with a single LLM call ($\sim$5 seconds, $\sim$\$0.01):

\begin{lstlisting}[language=yaml,basicstyle=\ttfamily\footnotesize]
# Quick analyze result for a new document
matches_schema: false
new_entities: [Acquisition, Partnership]
confidence: 0.85
reasoning: "Document contains M&A events not modeled in schema"
\end{lstlisting}

\subsection{Deep Extension Pipeline}
\label{app:extend:deep}

When sufficiency analysis indicates extension is needed, we invoke the full analysis pipeline.

\subsubsection{Analysis Phase}

Persistent extension reuses the same analyzers as initial schema generation (Section~\ref{app:schema}):
\begin{enumerate}
  \item \textbf{Two-pass LLM discovery:} Deep + validation passes to extract entities and relationships from the new data.
  \item \textbf{Semantic enrichment:} Detect additional entity types from data values (not field names alone).
  \item \textbf{Structural analysis:} Detect identity keys, foreign keys, and JSON paths -- all without LLM involvement.
\end{enumerate}

\noindent The output has the same form as the initial schema. Query-scoped extension uses a lighter single-call analysis (Section~\ref{app:extend:temp}).

\subsection{Diff-Based Merge}
\label{app:extend:merge}

We compute a schema diff to identify additions without modifying existing definitions.

\begin{algorithm}[t]
\caption{Schema Diff Computation}
\label{alg:diff}
\begin{algorithmic}[1]
\Procedure{ComputeDiff}{$\mathit{analysis}, \Sigma$}
  \State $\mathit{diff} \gets \{\mathit{new\_entities}: [], \mathit{new\_attrs}: \{\}, \mathit{new\_rels}: []\}$
  \State $E_{\mathit{existing}} \gets \Call{GetEntities}{\Sigma}$
  \For{$e \in \mathit{analysis}.\mathit{entities}$}
    \If{$e.\mathit{type} \notin E_{\mathit{existing}}$}
      \State $\mathit{diff}.\mathit{new\_entities}.\Call{append}{e}$
    \Else
      \State $A_{\mathit{new}} \gets e.\mathit{attrs} \setminus E_{\mathit{existing}}[e.\mathit{type}].\mathit{attrs}$
      \If{$|A_{\mathit{new}}| > 0$}
        \State $\mathit{diff}.\mathit{new\_attrs}[e.\mathit{type}] \gets A_{\mathit{new}}$
      \EndIf
    \EndIf
  \EndFor
  \State \textit{// Similar logic for relationships and structural intelligence}
  \State \Return $\mathit{diff}$
\EndProcedure
\end{algorithmic}
\end{algorithm}

\paragraph{Merge rules.}
\begin{itemize}
  \item New entity types are appended (never replace existing).
  \item New attributes extend existing entity definitions.
  \item Conflicting types are flagged; existing definition takes precedence.
  \item Version metadata tracks each extension for audit.
\end{itemize}

In query-scoped extension, only additions with LLM-assigned confidence $\geq 0.7$ are merged; persistent extension merges all additions that survive the diff, since the full analysis pipeline already provides high-quality outputs. Both modes enforce the monotonicity invariant: existing definitions are never removed or modified.

\subsection{Persistent Extension}
\label{app:extend:persistent}

Persistent extension writes the updated schema to disk with version tracking.

\subsubsection{Version Management}

Each extension increments the schema version and logs changes:

\begin{lstlisting}[language=yaml,basicstyle=\ttfamily\footnotesize]
# Schema header after extension
schema_version: 2
extensions:
  - version: 2
    timestamp: "2026-01-26T14:30:00Z"
    added_entities: [Acquisition]
    added_relationships: [ACQUIRED]
    added_attributes:
      Company: [acquisition_count]
    structural_changes:
      identity_keys: [Acquisition.deal_id]
      foreign_keys: [acquirer_qid -> entity_id]
\end{lstlisting}

\subsubsection{Extension Log}

A separate JSON log provides detailed audit trail:

\begin{lstlisting}[language=yaml,basicstyle=\ttfamily\footnotesize]
{
  "timestamp": "2026-01-26T14:30:00Z",
  "source_document": "new_acquisitions.jsonl"
}
\end{lstlisting}

\subsection{Query-Scoped Extension}
\label{app:extend:temp}

Query-scoped extension modifies an in-memory copy without persisting changes.

\subsubsection{Context Manager Pattern}

We implement automatic cleanup via Python context managers:

\begin{lstlisting}[language=python,basicstyle=\ttfamily\footnotesize]
# Automatic revert after query processing
with extender.query_context(query) as extended_schema:
    entities = extract_entities(data, extended_schema)
    answer = retrieve_and_answer(query, entities)
# Schema automatically reverted here
\end{lstlisting}

\subsubsection{Extension Lifecycle}

\begin{algorithm}[t]
\caption{Query-Scoped Extension}
\label{alg:query-extend}
\begin{algorithmic}[1]
\Procedure{ExtendForQuery}{$Q, \Sigma_{\mathit{base}}$}
  \State $\mathit{analysis} \gets \Call{AnalyzeQueryRequirements}{Q, \Sigma_{\mathit{base}}}$
  \If{$\neg \mathit{analysis}.\mathit{needs\_extension}$}
    \State \Return $\Sigma_{\mathit{base}}$
  \EndIf
  \State $\Sigma' \gets \Call{DeepCopy}{\Sigma_{\mathit{base}}}$
  \For{$e \in \mathit{analysis}.\mathit{required\_entities}$}
    \If{$e.\mathit{confidence} \geq 0.7$}
      \State $\Call{AddEntity}{\Sigma', e}$
    \EndIf
  \EndFor
  \State \Return $\Sigma'$
\EndProcedure
\Procedure{RevertToBase}{}
  \State $\Sigma' \gets \Call{DeepCopy}{\Sigma_{\mathit{base}}}$ \Comment{Discard extensions}
\EndProcedure
\end{algorithmic}
\end{algorithm}

The extended schema is used for the current query's retrieval and then automatically reverted via the context manager.
Optional snapshots can be saved for manual review and later promotion to a persistent schema.

\subsubsection{Snapshot for Manual Promotion}

Query-scoped extensions can optionally be saved as snapshots for later review:

\begin{lstlisting}[language=python,basicstyle=\ttfamily\footnotesize]
result = extender.extend_for_query(query, save_snapshot=True)
# Snapshot saved to: snapshots/schema_20260126_143000.yaml

# Later, manually review and load as persistent schema if useful
\end{lstlisting}

Snapshots are YAML files written to a \texttt{snapshots/} directory. Promotion to a persistent schema requires manual review and loading -- an intentional design choice to prevent untested query-driven entities from silently entering the production schema.





\section{Schema-Guided QA Agent: Technical Details}
\label{app:agent}

This appendix details the query-time agent described in Section~\ref{sec:agent-main}.
Given a query $Q$, a schema $\Sigma'$ (possibly query-extended), structured data stores (DataFrames loaded via the schema's source catalog), and a dense text index (FAISS over embeddings), the agent performs \emph{schema-conditioned query analysis}, \emph{tool routing}, \emph{reliability gating}, and \emph{answer synthesis}.

\subsection{Tools and I/O Contracts}
\label{app:agent:tools}

The agent exposes a modular tool set with explicit I/O contracts; Section~\ref{app:agent:routing} gives the routing policy over these tools.

\noindent\textbf{\textsc{SchemaLookup}} (deterministic structured lookup).
Input: entity type, key/value constraints, and an optional return field -- all derived from $\Sigma'$.
Execution: loads the relevant source DataFrame (identified by $\Sigma'$'s source catalog), applies exact-then-fuzzy constraint matching, and extracts the return field.  Supports nested JSON-in-CSV fields via automatic parsing.
Output: matching attribute values and per-constraint match scores.

\noindent\textbf{\textsc{VectorRAG}} (dense retrieval over unstructured evidence).
Input: query string and optional entity/temporal filters (e.g., ticker, quarter).
Execution: encodes the query with the index's embedding model, applies FAISS \texttt{IDSelector}-based pre-filtering when entity constraints are available, and optionally reranks candidates with a cross-encoder (BAAI/bge-reranker-large).
Output: top-$k$ text chunks with similarity scores.

\noindent\textbf{\textsc{GraphTraverse}} (multi-hop traversal over the KG and dataset-provided links).
Input: a start entity and a relationship pattern, both derived from $\Sigma'$.
Execution: queries the Neo4j knowledge graph built during ingestion (Section~\ref{sec:ingest-main}), following the typed relationships and cross-source edges declared in $\Sigma'$. It also follows links supplied with the benchmarks, which exist even without the KG (e.g., Wikipedia links in HybridQA, Wikidata entity lookups in BlendQA).
Output: matching entities and paths.

\noindent\textbf{\textsc{ExternalSearch}} (web search for OOD queries).
Input: query string.
Output: web snippets for queries that fall outside the schema's coverage (see Section~\ref{app:agent:ood}). Disabled in all benchmark runs (Section~\ref{app:agent:routing}).

\subsection{Schema-Conditioned Query Analysis}
\label{app:agent:analysis}

The agent extracts (i) candidate entity types from $\Sigma'$ and (ii) an expected answer type.
In our implementation, entity-type selection is performed either via lightweight heuristics (e.g., table-column typing in HybridQA) or via constrained LLM classification over the schema vocabulary (e.g., ComplexTR).
Answer type is predicted as one of \{\texttt{string}, \texttt{number}, \texttt{date}, \texttt{entity list}\}, which guides tool selection and output validation.

\subsection{OOD Detection}
\label{app:agent:ood}

The two-stage reliability gating described in Section~\ref{sec:agent-main} is implemented as follows.

\paragraph{Pre-retrieval (Stage~1).}
Before any tool is invoked, the agent checks:
\begin{enumerate}
  \item \textbf{Temporal OOD}: keywords indicating real-time information needs (``current'', ``latest'', ``today'') or year references beyond $\Sigma'$'s temporal range.
  \item \textbf{Entity OOD}: extracted entity mentions (e.g., tickers) that do not appear in $\Sigma'$'s known-value dictionaries.
\end{enumerate}
If either check fires, the query is flagged OOD and routed directly to \textsc{ExternalSearch}.

\paragraph{Post-retrieval (Stage~2).}
After tool execution, the agent checks:
\begin{enumerate}
  \item \textbf{Retrieval confidence}: if the maximum similarity score across retrieved passages falls below a threshold (default 0.3), the result is flagged as low-confidence.
  \item \textbf{Answer grounding}: if the generated answer contains uncertainty phrases (e.g., ``I cannot determine''), the result is flagged.
\end{enumerate}
Flagged queries trigger a fallback to \textsc{ExternalSearch} and confidence is discounted.

\subsection{Routing Policy}
\label{app:agent:routing}

Routing is \emph{schema-driven}: the agent prefers deterministic structure when the schema exposes typed, discrete-valued fields.
The policy follows a preference order:

\begin{itemize}
  \item If $Q$'s constraints map to typed columns with discrete values in $\Sigma'$ (e.g., identifier fields, categorical attributes, temporal keys), invoke \textsc{SchemaLookup} on the source identified by $\Sigma'$'s entity-to-source mapping.
  \item If $Q$ requires descriptive evidence or $\Sigma'$ indicates unstructured fields are relevant, invoke \textsc{VectorRAG} with entity/temporal pre-filtering derived from $Q$'s constraints.
  \item If $Q$ is multi-hop or asks how entities are related, invoke \textsc{GraphTraverse} over the KG, following the relationships and cross-source edges declared in $\Sigma'$.
  \item If the pre-retrieval OOD gate fires, route to \textsc{ExternalSearch}.
  \item If \textsc{SchemaLookup} returns empty, the agent retries with \textsc{VectorRAG}; if both local tools are exhausted, it escalates to \textsc{ExternalSearch}.
\end{itemize}

\noindent In all benchmark runs \textsc{ExternalSearch} is disabled, so these queries are routed to abstention.

\subsection{Plan--Execute--Retry Loop}
\label{app:agent:loop}

Algorithm~\ref{alg:agent} summarizes the agent loop, including the retry mechanism with failure summaries described in Section~\ref{sec:agent-main}.

\begin{algorithm}[t]
\caption{Schema-Guided QA Agent with Reliability Gating}
\label{alg:agent}
\begin{algorithmic}[1]
\Procedure{Answer}{$Q, \Sigma, \mathcal{B}$} \Comment{$\mathcal{B}$: data stores and text index}
  \State $\Sigma' \gets$ \Call{MaybeExtendSchema}{$Q,\Sigma$}
  \If{\Call{PreRetrievalOOD}{$Q,\Sigma'$}} \Return \Call{ExternalSearch}{$Q$}
  \EndIf
  \State $(T, A) \gets$ \Call{AnalyzeQuery}{$Q,\Sigma'$} \Comment{entity types, answer type}
  \State $\mathit{failure} \gets \textsc{None}$
  \For{$r = 0$ \textbf{to} $R_{\max}$}
    \State $\mathcal{P} \gets$ \Call{RouteTools}{$Q,T,A,\Sigma',\mathit{failure}$}
    \State $\mathcal{E} \gets \bigcup_{p \in \mathcal{P}}$ \Call{Execute}{$p, Q, \Sigma', \mathcal{B}$}
    \State $\hat{y} \gets$ \Call{ExtractAnswer}{$Q, \mathcal{E}$}
    \State $(\hat{y}_g, g) \gets$ \Call{GroundAnswer}{$\hat{y}, \mathcal{E}$} \Comment{grounded answers, type}
    \State $c \gets c_{\text{link}} \cdot c_{\text{node}} \cdot c_{\text{ground}}(g)$
    \If{$\hat{y}_g \neq \emptyset$} \Return $(\hat{y}_g, c)$
    \EndIf
    \State $\mathit{failure} \gets$ \Call{BuildFailureSummary}{$\mathcal{E}$}
  \EndFor
  \State \Return $(\emptyset, 0)$ \Comment{abstain after $R_{\max}$ retries}
\EndProcedure
\end{algorithmic}
\end{algorithm}

\paragraph{Confidence decomposition.}
The composite confidence $c = c_{\text{link}} \cdot c_{\text{node}} \cdot c_{\text{ground}}$ decomposes as follows.
$c_{\text{link}}$ captures path confidence (1.0 for single-tool plans, discounted for multi-tool compositions).
$c_{\text{node}} = \min_i m_i$ is the minimum constraint-match score across all query constraints, where $m_i = 1.0$ for exact match and $0.8$ for substring containment match.
$c_{\text{ground}}$ reflects evidence support: 1.0 if answers match structured candidates exactly, 0.9 for passage-supported answers, 0.7 for partial passage support, and 0.0 when no evidence supports the answer.

\paragraph{Failure summary and retry.}
When no grounded answer is found, the agent builds a structured failure summary identifying: (i)~which tools were tried and returned empty, and (ii)~which tools were \emph{not} tried and should be included in the next attempt.
For example, if \textsc{SchemaLookup} returned no candidates and \textsc{VectorRAG} was not invoked, the failure summary mandates \textsc{VectorRAG} in the retry.  If all local tools are exhausted, \textsc{ExternalSearch} is escalated.

\subsection{Execution Traces and Diagnostics}
\label{app:agent:traces}

\paragraph{Fallback.} When the LLM is unavailable, the agent returns schema-lookup candidates directly ($c{=}1.0$ if grounded). Two-stage gating with retry provides principled abstention on OOD queries while maximizing recall on in-distribution ones.

\paragraph{Traces.} For each query we record: invoked tools, tool outputs (e.g., number of structured matches, retrieved passages, match scores), and latency.
These traces support routing diagnostics such as tool usage rates, OOD detection rates, and tool-specific failure modes, complementing end-to-end QA metrics.

\section{LLM Prompts}
\label{app:prompts}

This section reproduces the core prompt templates used in schema generation
(\S\ref{app:schema}), KG ingestion (\S\ref{app:ingest}), and the QA agent
(\S\ref{app:agent}).
Each prompt is shown in condensed form with variable placeholders
(\texttt{\{...\}}) indicating inputs injected at runtime.
Output-format sections show illustrative instantiations; actual entity types,
relationship types, and IDs are populated from the discovered schema at
runtime.

\subsection{Schema Discovery Prompt (Deep Discovery)}
\label{app:prompt:discovery}

This prompt is the first LLM call in the two-pass schema generation
pipeline (\S\ref{app:schema:llm}).
It receives the field catalog produced by the structural analyzer and a
set of matched samples, and asks the model to propose entity types,
relationships, temporal configuration, and extraction methods.
At runtime, format-adaptive interpretation rules
(e.g., table-column entity discovery for tabular sources, nested-structure
handling for JSON) are automatically selected based on the detected source
format; these are omitted here for space.

\begin{tcolorbox}[colback=gray!5,colframe=gray!75,title=Deep-Discovery Prompt Template,boxsep=2pt,left=3pt,right=3pt,top=3pt,bottom=3pt]
\footnotesize\ttfamily
You are an expert schema discovery agent.\\
Your job is to infer an initial single-source schema
by OBSERVING ONLY:\\
~~1) the FIELD CATALOG (authoritative list of fields), and\\
~~2) the SAMPLE DATA provided (may include multiple samples).

\smallskip
DO NOT invent fields, entities, or relationships that are
not evidenced in the samples.\\
PRIORITIZE COMPLETENESS: model ALL entity types and
relationships across all samples provided.

\smallskip
\textbf{Dataset:} \{description\}\\
\textbf{Source Format:} \{format\}~~(JSON / CSV / Parquet / Text / Hybrid)\\
\textbf{Total Records:} \{num\_records\}

\smallskip
\textbf{FIELD CATALOG (authoritative):}\\
\{catalog\_json\}

\smallskip
\textbf{SAMPLE DATA (read ALL carefully; only evidence):}\\
\{sample\_str\}

\smallskip
\textbf{=== STEP A  --  OBSERVE (EVIDENCE FIRST) ===}\\
List concrete patterns: identifiers/keys, repeated labels,
turn-taking markers, date/time expressions, nested structures.\\
For each observation capture field\_id + short snippet.

\smallskip
\textbf{=== STEP B  --  ENTITIES (COMPREHENSIVE DISCOVERY) ===}\\
Discover TWO kinds of entities:\\
~~1. INSTANCE-BASED: same specific instance appears multiple
times (e.g., same person, organization, identifier).\\
~~2. CONCEPTUAL / CATEGORICAL: different instances of the same
concept type (e.g., FinancialMetric, Product, TimePeriod).

\smallskip
For each entity provide: name, description, anchor
(primary key or natural key), attributes (field\_id, name,
type\_guess), extraction method (field $|$ nested\_field $|$
pattern $|$ llm), examples, and confidence in [0,1].\\
Include a SourceRecord entity for provenance tracking.

\smallskip
\textbf{=== STEP C  --  RELATIONSHIPS (COMPREHENSIVE) ===}\\
Propose ALL relationships evidenced in the data:\\
~~- STRUCTURAL (foreign keys, co-occurrence in records)\\
~~- SEMANTIC (text spans, possessive/attributive patterns)\\
~~- IMPLICIT (metric $\rightarrow$ time period when co-occurring)

\smallskip
For each: type (UPPER\_SNAKE), source/target entities,
cardinality, evidence snippet, extraction strategy
(join\_on\_field $|$ co\_occurrence $|$ within\_text\_span $|$
semantic\_extraction), and confidence.

\smallskip
\textbf{=== STEP D  --  TEMPORAL CONFIG ===}\\
Enable temporal ONLY if sample shows time-bound
relationships.\\
Choose: edge\_properties $|$ reified\_nodes $|$
date\_nodes $|$ none.

\smallskip
\textbf{OUTPUT:} strict JSON with keys \{observations, entities,
relationships, temporal, notes\}.
\end{tcolorbox}

\subsection{Schema Validation Prompt}
\label{app:prompt:validation}

The second LLM call (\S\ref{app:schema:llm}) reviews additional
(diverse) samples against the existing schema to detect gaps.
It preserves the existing schema and proposes \emph{additions only}.

\begin{tcolorbox}[colback=gray!5,colframe=gray!75,title=Validation \& Extension Prompt Template,boxsep=2pt,left=3pt,right=3pt,top=3pt,bottom=3pt]
\footnotesize\ttfamily
You are a schema validation + extension agent.\\
Review ADDITIONAL SAMPLES and the FIELD CATALOG
to find NEW patterns, entities, or relationships
that complement the existing schema.

\smallskip
\textbf{GUIDELINES:}\\
1) Preserve existing schema by default.\\
2) Add NEW entities for any distinct uncaptured concepts.\\
3) Add NEW relationships for any clear connections.\\
4) Add NEW attributes for existing entities if found.\\
5) Note issues but do not remove existing elements.

\smallskip
\textbf{FIELD CATALOG (authoritative):}\\
\{catalog\_summary\}

\smallskip
\textbf{EXISTING SCHEMA (authoritative; must be preserved):}\\
~~Entities: \{entity\_summary\}\\
~~Relationships: \{relationship\_summary\}\\
~~Temporal: \{temporal\_info\}

\smallskip
\textbf{ADDITIONAL SAMPLES (only for gap finding):}\\
\{samples\_str\}

\smallskip
\textbf{WHAT TO LOOK FOR:}\\
Step 1  --  Check field catalog for uncovered fields that
represent distinct concepts.\\
Step 2  --  Check sample data:\\
~~Add entity if: distinct queryable concept, conceptual
category, or recurring multi-context concept.\\
~~Add relationship if: structural or semantic connection
observed, including LLM-extractable ones.

\smallskip
\textbf{EVIDENCE RULES:}\\
Every addition requires: evidence snippet ($\leq$25 words),
source field\_id, extraction method (field $|$ nested\_field $|$
pattern $|$ llm), confidence in [0,1].

\smallskip
\textbf{OUTPUT:} strict JSON with keys \{added\_entities,
added\_relationships, added\_attributes, temporal\_changes,
issues, notes\}.
\end{tcolorbox}

\subsection{Query-Driven Schema Extension Prompt}
\label{app:prompt:query-ext}

When the QA agent encounters a query whose required concepts are not
covered by the current schema, the
\texttt{QueryDrivenSchemaExtender} invokes this analysis prompt
to determine whether on-the-fly extension is warranted
(\S\ref{app:extend:temp}).

\begin{tcolorbox}[colback=gray!5,colframe=gray!75,title=Query-Driven Schema Analysis Prompt,boxsep=2pt,left=3pt,right=3pt,top=3pt,bottom=3pt]
\footnotesize\ttfamily
Analyze this user query to determine if we need
additional entity types or relationships beyond
the current knowledge graph schema.

\smallskip
\textbf{Query:} "\{query\}"

\smallskip
\textbf{Existing entities:} \{existing\_entities\}\\
\textbf{Existing relationships:} \{existing\_relationships\}

\smallskip
\textbf{Task:}\\
1. Identify entity types mentioned or implied.\\
2. Determine if any are MISSING from existing entities.\\
3. Identify any relationships needed that are MISSING.\\
4. For each missing item, provide:\\
~~~- Type name\\
~~~- Confidence (0.0--1.0)\\
~~~- Reason why needed\\
~~~- Example instances from the query\\
~~~- Attributes to track

\smallskip
\textbf{Return JSON:}\\
\{~"needs\_extension": true/false,\\
~~"required\_entities": [\\
~~~~\{"type": "...", "confidence": 0.95,\\
~~~~~"reason": "...", "examples": [...],\\
~~~~~"attributes": [...]\}~],\\
~~"required\_relationships": [\\
~~~~\{"type": "...", "source\_entity": "...",\\
~~~~~"target\_entity": "...", "confidence": 0.9,\\
~~~~~"reason": "..."\}~],\\
~~"reasoning": "..."~\}

\smallskip
Important: Only suggest extensions if truly needed.\\
Return empty lists if existing schema is sufficient.
\end{tcolorbox}

\subsection{QA Agent: Pre-Retrieval Prompt}
\label{app:prompt:preretrieve}

The pre-retrieval prompt performs schema-conditioned query analysis and
tool selection (\S\ref{app:agent:analysis}, \S\ref{app:agent:routing}).
It receives the full schema and the user query, and outputs a tool
invocation plan.  An OOD gate (\S\ref{app:agent:ood}) is applied before
tool execution.

\begin{tcolorbox}[colback=gray!5,colframe=gray!75,title=Pre-Retrieval: Schema Analysis \& Tool Selection,boxsep=2pt,left=3pt,right=3pt,top=3pt,bottom=3pt]
\footnotesize\ttfamily
You are a schema-guided question-answering agent. Analyze the
question against the schema to determine which tools to invoke.

\smallskip
\textbf{=== SCHEMA (YAML) ===}\\
\{schema\_yaml\}

\smallskip
\textbf{=== AVAILABLE TOOLS ===}

\textbf{SchemaLookup(entity\_type, constraints, return\_field)}\\
~~Deterministic lookup on structured data sources.\\
~~Use for: typed attribute queries, identifier lookups.

\textbf{VectorRAG(query, filters, top\_k)}\\
~~Dense retrieval over unstructured text passages.\\
~~Use for: descriptive evidence, open-ended questions.

\textbf{GraphTraverse(start\_entity, pattern)}\\
~~Follow links in the knowledge graph.\\
~~Use for: multi-hop reasoning, relationship traversal.

\textbf{ExternalSearch(query)}\\
~~Web search fallback for out-of-distribution queries.\\
~~Use for: queries outside schema coverage.

\smallskip
\textbf{=== QUESTION ===}\\
\{query\}

\smallskip
\textbf{=== INSTRUCTIONS ===}\\
1. Parse the schema to identify relevant entity types,\\
~~~relationships, and field semantics.\\
2. Check for OOD signals: temporal keywords (``current'',\\
~~~``latest''), entity mentions absent from schema.\\
~~~If OOD, route directly to ExternalSearch.\\
3. Select tools based on query type:\\
~~~- Typed, discrete-valued constraints $\rightarrow$ SchemaLookup\\
~~~- Descriptive / unstructured evidence $\rightarrow$ VectorRAG\\
~~~- Multi-hop or relationship queries $\rightarrow$ GraphTraverse\\
~~~- Out-of-distribution $\rightarrow$ ExternalSearch\\
4. Generate tool calls with schema-derived parameters.

\smallskip
\textbf{OUTPUT:}\\
\{tool\_calls\}
\end{tcolorbox}

\newpage  

\section{Additional Analyses}
\label{app:additional}

\subsection{Judge Robustness}
\label{app:multijudge}

To check whether LLM-J depends on the choice of judge, we rescored BlendQA answers from the GPT-4.1 schema-guided system with three judges from different model families, using the same binary prompt (Table~\ref{tab:multijudge}).

\begin{table}[h]
\centering
\small
\begin{tabular}{lc}
\toprule
\textbf{Judge} & \textbf{LLM-J (\%)} \\
\midrule
GPT-4.1 & 72.3\\
Llama 3.3 70B & 72.8 \\
Claude Haiku 4.5 & 70.0 \\
\midrule
Pairwise agreement (\%) & 92.0 \\
\bottomrule
\end{tabular}
\caption{LLM-J on BlendQA scored by three judges. }
\label{tab:multijudge}
\end{table}

\noindent The three judges give accuracies within 2.8 points of each other. Claude Haiku~4.5, the judge used in the main results, is the strictest, 2.3 points below GPT-4.1.

\subsection{Cross-Model Evaluation on TAT-QA}
\label{app:crossmodel-tatqa}

To test whether the gains on \textsc{TAT-QA} depend on the GPT-4.1 backbone, we rerun RAG, ReAct, and the Schema-Guided system with two other backbones: Claude Haiku~4.5 and Llama~3.1~70B (the Llama~3.3~70B endpoint was unavailable for these runs). Schema, KG, sources, and exclusions are unchanged; only the backbone LLM differs. EM and F1 use the official TAT-QA scorer, and LJ is scored by Claude Haiku~4.5 for both backbones (Table~\ref{tab:crossmodel-tatqa}).

\begin{table}[h]
\centering
\small
\setlength{\tabcolsep}{4pt}
\begin{tabular}{llcccc}
\toprule
\textbf{Backbone} & \textbf{Method} & \textbf{EM} & \textbf{F1} & \textbf{BLEU} & \textbf{LJ} \\
\midrule
Claude      & RAG   & 27.2 & 37.0 & 43.2 & 49.6 \\
Haiku 4.5   & ReAct & 41.7 & 52.1 & 60.5 & 66.7 \\
            & Ours  & \textbf{81.1} & \textbf{86.7} & \textbf{87.3} & \textbf{89.4} \\
\midrule
Llama       & RAG   & 27.2 & 35.3 & 35.9 & 49.4 \\
3.1 70B     & ReAct & 35.6 & 41.7 & 43.6 & 63.6 \\
            & Ours  & \textbf{71.1} & \textbf{76.2} & \textbf{76.0} & \textbf{79.4} \\
\bottomrule
\end{tabular}
\caption{Cross-model results on \textsc{TAT-QA} (\%). EM/F1 use the official TAT-QA scorer; LJ is judged by Claude Haiku~4.5. Bold marks the best result for each backbone.}
\label{tab:crossmodel-tatqa}
\end{table}

\noindent The Schema-Guided system beats both baselines with both backbones. Over ReAct it gains $+$39.4 EM / $+$22.7 LJ with Claude Haiku~4.5 and $+$35.5 EM / $+$15.8 LJ with Llama~3.1~70B. Over RAG, the EM gains are larger still ($+$53.9 and $+$43.9). The ranking RAG $<$ ReAct $<$ Ours holds for both backbones, so the gains on \textsc{TAT-QA} are not specific to GPT-4.1.

\subsection{Run-to-Run Variance}
\label{app:variance}

API-served models are not fully deterministic at temperature~$=0$, so we ran the Schema-Guided system on \textsc{BlendQA} three times (Table~\ref{tab:variance}).

\begin{table}[h]
\centering
\small
\resizebox{\columnwidth}{!}{%
\begin{tabular}{lcccc}
\toprule
 & \textbf{EM} & \textbf{F1} & \textbf{BLEU} & \textbf{LJ} \\
\midrule
Schema-Guided & 29.4 $\pm$ 0.3 & 48.8 $\pm$ 0.2 & 54.1 $\pm$ 0.3 & 70.0 $\pm$ 0.3 \\
\bottomrule
\end{tabular}%
}
\caption{Schema-Guided results on \textsc{BlendQA} over three runs (mean $\pm$ standard deviation; GPT-4.1, temperature~$=0$).}
\label{tab:variance}
\end{table}

\subsection{Is the Graph Doing the Work?}
\label{app:kg-flips}

High graph \emph{usage} does not by itself show that the graph is \emph{necessary}: the router invokes \textsc{GraphTraverse} in 96\% of correctly answered queries because traversal is cheap (4.6\,ms vs.\ 584\,ms for vector retrieval; Table~\ref{tab:online_metrics}). To measure necessity, we compare the full system with the w/o-KG ablation question by question and count answers that flip in each direction (Table~\ref{tab:kg-flips}).

\begin{table}[h]
\centering
\small
\setlength{\tabcolsep}{4pt}
\begin{tabular}{lccc}
\toprule
\textbf{Dataset} & \textbf{Fixed by KG} & \textbf{Broken by KG} & \textbf{$\Delta$EM} \\
\midrule
\textsc{BlendQA}   & 10.2\% & 2.2\% & $+$8.0 \\
\textsc{HybridQA}  & 6.4\% & 5.9\% & $+$0.5 \\
\textsc{TAT-QA}    & 3.5\% & 0.9\% & $+$2.6 \\
\textsc{ComplexTR} & 8.8\% & 3.1\% & $+$5.7 \\
\bottomrule
\end{tabular}
\caption{Per-question flips between the full system and the w/o-KG ablation. \emph{Fixed}: wrong without the KG, right with it; \emph{Broken}: the reverse. $\Delta$EM is from Table~\ref{tab:qa_all_metrics}; when flips are counted on EM, Fixed $-$ Broken equals $\Delta$EM.}
\label{tab:kg-flips}
\end{table}

\noindent On \textsc{BlendQA}, the fixed answers are concentrated in cross-source multi-hop questions. On \textsc{HybridQA} and \textsc{TAT-QA}, removing the KG changes EM little, consistent with Section~\ref{sec:ablations}: there, schema-guided structured lookup carries most of the load. The graph thus drives the cross-source multi-hop gains, while structured access drives the table-centric ones; the same fallback redundancy keeps partial schema realization (SchUtil 50\% on \textsc{TAT-QA}) from reducing answerability.

\subsection{Comparison with the Closest Systems}
\label{app:closest}

Table~\ref{tab:closest} compares our system with the three closest systems on what they assume as input and what they automate.

\begin{table*}[t]
\centering
\small
\setlength{\tabcolsep}{4pt}
\begin{tabularx}{\linewidth}{@{}>{\raggedright\arraybackslash}p{2.6cm}>{\raggedright\arraybackslash}X>{\raggedright\arraybackslash}X*{3}{>{\centering\arraybackslash}p{1.5cm}}@{}}
\toprule
\textbf{System} & \textbf{Assumed input / substrate} & \textbf{What is built automatically} & \textbf{Typed schema} & \textbf{Cross-source keys} & \textbf{Query-time extension} \\
\midrule
AtomR\newline\cite{xin2025atomr} & Existing heterogeneous knowledge sources with retrieval or query interfaces & Atomic question decomposition and reasoning plan & \xmark & \xmark & \xmark \\
HippoRAG~2\newline\cite{gutierrez2025rag} & Text passages & Open (schema-free) triple graph and passage index & \xmark & \xmark & \xmark \\
GraphRAG\newline\cite{edge2024graphrag} & Text corpus & LLM-extracted entity graph and community summaries & \xmark & \xmark & \xmark \\
Ours & Raw tables, JSON, and text & Typed schema, identity/foreign keys, provenance KG, schema-conditioned routing & \cmark & \cmark & \cmark \\
\bottomrule
\end{tabularx}
\caption{Assumptions and automated components of the closest systems. }
\label{tab:closest}
\end{table*}

\section{Adapted GraphRAG Baseline}
\label{app:graphrag}

GraphRAG~\cite{edge2024graphrag} builds community-summarized graphs over text corpora and retrieves via community reports, targeting homogeneous text summarization rather than multi-source schema-guided QA. Applying it to our setting requires serializing heterogeneous tables, JSON, and structured records into textual evidence units -- design choices absent from GraphRAG's original formulation. We therefore report this comparison as an appendix diagnostic rather than in the main controlled baseline block. The adapted pipeline uses the same GPT-4.1 backbone, dataset-provided sources, exclusion manifest, and evaluation harness as the controlled baselines, but does not use our induced schema, foreign-key paths, provenance graph, or schema-conditioned router. We augment the standard GraphRAG indexing pipeline with three dataset-aware preprocessing steps: (1) metadata saturation, where key identifiers are re-injected every ${\sim}$800 characters so they survive chunking and remain available at retrieval time; (2) structured-data serialization, where tables are converted to pipe-delimited markdown with explicit header rows and paragraphs are appended as contiguous text blocks, preserving the tabular-textual duality of the source; and (3) domain-tuned entity schemas, where entity types are customized per dataset (e.g., adding metric, segment, and date for financial QA) rather than relying on generic defaults. At retrieval time, we shift weight toward text units and increase the entity/relationship fan-out (top-$k{=}30$) relative to GraphRAG defaults, favoring fine-grained lexical evidence over community summaries for fact-centric QA. GraphRAG scores highest on ComplexTR (45.2 EM), whose corpus consists of homogeneous textual fact sentences from a single source -- the setting GraphRAG was designed for -- but still trails our system by 29.3 EM there. The gap is largest on TAT-QA (55.7 EM), where tables and passages must be jointly queried. HybridQA is omitted because adapting GraphRAG to its 79K-document corpus was computationally prohibitive for this diagnostic assessment.

\begin{table}[t]
\centering
\small
\setlength{\tabcolsep}{4pt}
\begin{tabular}{llcccc}
\toprule
\textbf{Dataset} & \textbf{Method} & \textbf{EM} & \textbf{F1} & \textbf{BLEU} & \textbf{LJ} \\
\midrule
\textsc{BlendQA} & GraphRAG & 18.5 & 32.7 & 34.9 & 55.1 \\
 & Ours & \textbf{29.3} & \textbf{48.9} & \textbf{54.1} & \textbf{70.0} \\
\midrule
\textsc{TAT-QA} & GraphRAG & 18.6 & 24.9 & 22.7 & 42.6 \\
 & Ours & \textbf{74.3} & \textbf{82.6} & \textbf{68.6} & \textbf{91.8} \\
\midrule
\textsc{ComplexTR} & GraphRAG & 45.2 & 50.1 & 52.2 & 49.5 \\
 & Ours & \textbf{74.5} & \textbf{82.2} & \textbf{82.0} & \textbf{76.3} \\
\bottomrule
\end{tabular}
\caption{Adapted GraphRAG results on three benchmarks alongside our Schema-Guided results from Table~\ref{tab:qa_all_metrics}. GraphRAG uses dataset-aware serialization, domain-tuned entity schemas,
and the same GPT-4.1 backbone, sources, exclusions, and evaluation harness,
but not our induced schema, FK paths, provenance graph, or schema-conditioned router.}
\label{tab:graphrag}
\end{table}

\section{Use of AI Assistants}
\label{app:ai-assistants}

AI assistants were used during development for code debugging and polishing the manuscript's language. All scientific claims, experimental design, system architecture, results, and final text were devised and verified by the authors.